\documentclass[journal]{IEEEtran}
\usepackage{amsmath,amsfonts}

\usepackage{algorithm}
\usepackage{array}
\usepackage[caption=false,font=normalsize,labelfont=sf,textfont=sf]{subfig}
\usepackage{textcomp}
\usepackage{stfloats}
\usepackage{url}
\usepackage{verbatim}
\usepackage{graphicx}
\usepackage{cite}
\usepackage{amssymb}
\usepackage{extarrows}
\usepackage{multirow}
\usepackage{colortbl}
\usepackage{tabularx} 
\usepackage{arydshln}
\usepackage{wrapfig}
\usepackage{booktabs} \let\cline\cmidrule
\usepackage{algorithm}
\usepackage{algpseudocode}
\usepackage{setspace}
\usepackage{xcolor}
\usepackage{hyperref} 
\usepackage{enumerate}
\usepackage{makecell}
\usepackage{arydshln}
\usepackage{svg}
\usepackage{pifont}

\usepackage{hyperref}

\newcommand{\orcid}[1]{\href{https://orcid.org/#1}{\includegraphics[width=10pt]{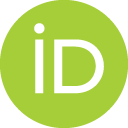}}}

\begin{document}

\title{DA-Cal: Towards Cross-Domain Calibration  in \\Semantic Segmentation}

\author{Wangkai Li \orcid{0009-0008-1776-8450}, Rui Sun\orcid{0000-0002-8009-4240}, Zhaoyang Li\orcid{0009-0000-6451-5378}, Yujia Chen\orcid{0009-0005-1533-9263}, Tianzhu Zhang  \orcid{0000-0003-1856-9564},~\IEEEmembership{Member,~IEEE}
        % <-this % stops a space

\thanks{This work was partially supported by the Science and Application Research Project (First Batch, Grant No. TW2-01-001) for the Tianwen-2 Mission under the Planetary Exploration Project.

The authors are with the Deep Space Exploration Laboratory,
School of Information Science and Technology, University of
Science and Technology of China (USTC), Hefei 230026, China. (e-mail: {lwklwk, issunrui, 
lizhaoyang, yujia\_chen}@mail.ustc.edu.cn;
{tzzhang}@ustc.edu.cn). Corresponding author: Tianzhu Zhang.}
}



\maketitle

\begin{abstract}
While existing unsupervised domain adaptation (UDA) methods greatly enhance target domain performance in semantic segmentation, they often neglect network calibration quality, resulting in misalignment between prediction confidence and actual accuracy---a significant risk in safety-critical applications.  Our key insight emerges from observing that performance degrades substantially when soft pseudo-labels replace hard pseudo-labels in cross-domain scenarios due to poor calibration, despite the theoretical equivalence of perfectly calibrated soft pseudo-labels to hard pseudo-labels. Based on this finding, we propose DA-Cal, a dedicated cross-domain calibration framework that transforms target domain calibration into soft pseudo-label optimization. DA-Cal introduces a Meta Temperature Network to generate pixel-level calibration parameters and employs bi-level optimization to establish the relationship between soft pseudo-labels and UDA supervision, while utilizing complementary domain-mixing strategies to prevent overfitting and reduce domain discrepancies. Experiments demonstrate that DA-Cal seamlessly integrates with existing self-training frameworks across multiple UDA segmentation benchmarks, significantly improving target domain calibration while delivering performance gains without inference overhead. 
\end{abstract}

    
\section{Introduction}
\label{sec:intro}

Semantic segmentation, a fundamental computer vision task that assigns semantic labels to image pixels, enables comprehensive scene understanding. Despite significant advances with  Deep Neural Networks (DNNs) \cite{long2015fully, chen2017rethinking, cheng2021per, cheng2022masked}, these methods typically experience performance degradation when applied to novel target datasets due to distribution shifts between domains. This domain gap presents substantial challenges for network generalization. To address this limitation, Unsupervised Domain Adaptation (UDA) \cite{tan2018survey} is introduced by transferring knowledge from labeled source domains to unlabeled target domains, improving target performance without requiring additional annotations.

\begin{figure}
    \centering
    \includegraphics[width=1\columnwidth]{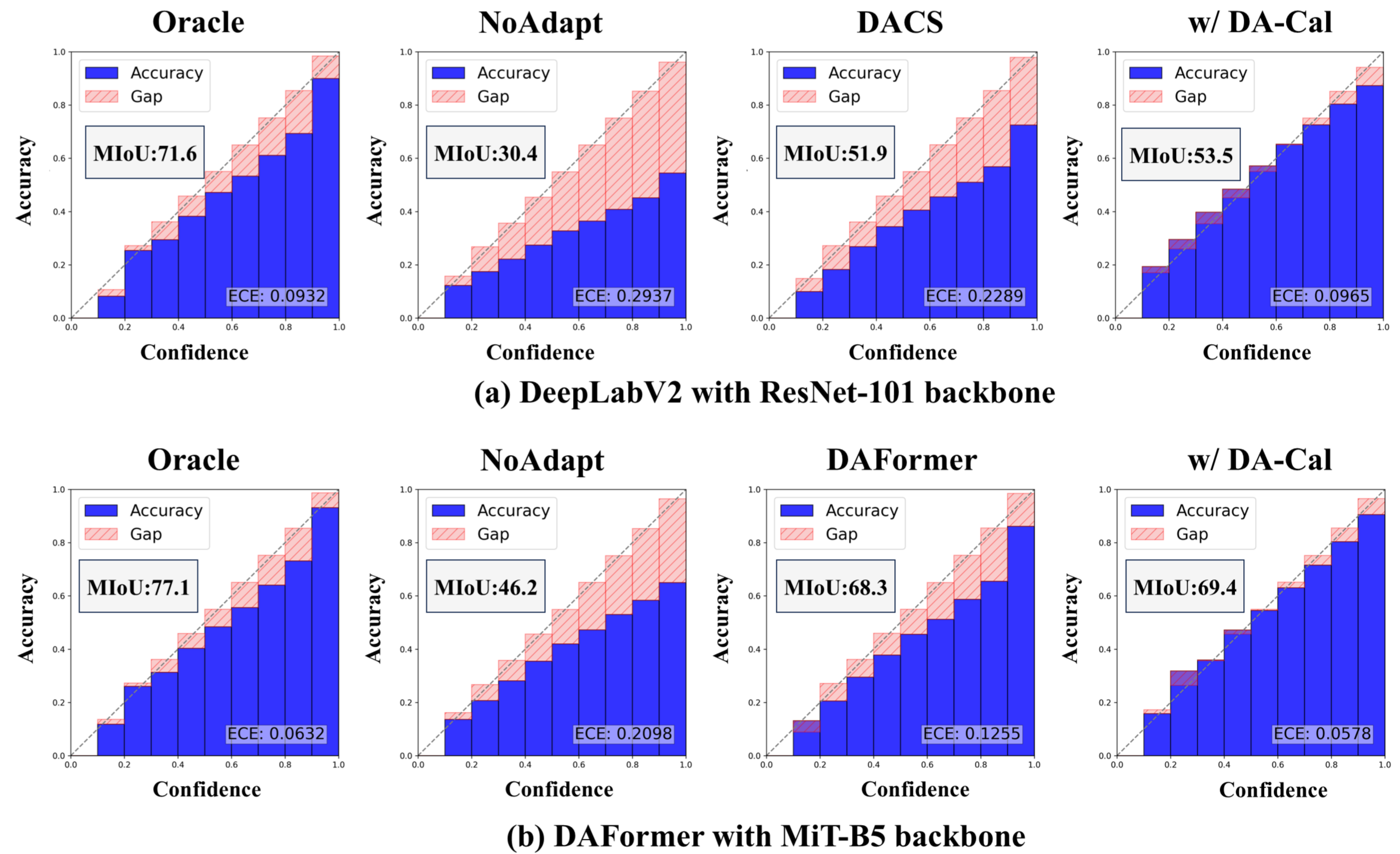}
    \caption{\textbf{ Reliability Diagrams \cite{niculescu2005predicting}.} We observe that neural networks not only degrade in performance under cross-domain settings but also exhibit poor calibration.  Although existing UDA methods, such as DACS \cite{tranheden2021dacs} and DAFormer \cite{hoyer2022daformer}, significantly improve target-domain performance, they still struggle to produce well-calibrated predictions. When integrated with our proposed DA-Cal, these methods achieve consistently improved calibration, leading to more reliable confidence estimates. }
    \label{fig1}
 % \vspace{-1em}
\end{figure}

However, for segmentation networks in real-world applications, reliable uncertainty estimation is often as critical as accuracy, particularly in safety-critical scenarios like autonomous driving \cite{bojarski2016end, ICLR2026_7f4c4786} and medical diagnosis \cite{esteva2017dermatologist, jiang2012calibrating, li2023maunet}. In these contexts, predictions must be both accurate and well-calibrated \cite{guo2017calibration, minderer2021revisiting}—where calibration refers to the alignment between network's confidence and its actual accuracy. Despite current UDA methods significantly narrowing performance gaps between domains, we observe that they fail to show corresponding improvements in calibration quality (Fig. \ref{fig1}). Moreover, state-of-the-art UDA approaches \cite{hoyer2023mic,yang2025micdrop} relying on self-training \cite{lee2013pseudo} may exacerbate confirmation bias, potentially degrading calibration further.

\IEEEpubidadjcol
% Current calibration methods fall into two categories:
% (1) Supervised IID calibration methods \cite{guo2017calibration, mozafari2018attended, ji2019bin} adjust network outputs through post-hoc techniques by learning additional hyperparameters like temperature scaling. These approaches depend on labeled in-distribution validation sets, making them unsuitable for UDA task due to: a) the absence of labeled target domain data, and b) severe domain shifts that render source-calibrated networks ineffective for the target domain \cite{wang2020transferable, de2023reliability}. 
% (2) Domain-shift calibration methods \cite{wang2020transferable, park2020calibrated} use importance weighting \cite{cortes2008sample} to establish relationships between target and source samples, addressing calibration under covariate shift. Additionally, \cite{hu2024pseudo} generates pseudo-target samples to simulate supervised calibration. Despite some progress, these methods are designed for classification tasks, and such sample-level modeling strategies are not guaranteed to work well for pixel-level semantic segmentation. \textit{How to achieve cross-domain calibration in semantic segmentation remains an unresolved challenge.}

% (Add to Intro, 3rd paragraph; modified text in red)
Current calibration methods fall into two categories:
(1) Supervised IID calibration methods \cite{guo2017calibration, mozafari2018attended, ji2019bin} adjust network outputs through post-hoc techniques by learning additional hyperparameters like temperature scaling. These approaches depend on labeled in-distribution validation sets, making them unsuitable for UDA tasks  due to: a) the absence of labeled target domain data, and b) severe domain shifts that render source-calibrated networks ineffective for the target domain \cite{wang2020transferable, de2023reliability}. 
{\color{black}Moreover, many segmentation-specific post-hoc calibrators (e.g., local/pixel-wise temperature scaling \cite{ding2021local, wang2023calibrating}) are typically fitted with labeled IID validation data; when fitted on the source domain, their calibration parameters often do not transfer well to the target domain under UDA, while fitting pixel-wise temperatures on unlabeled target data can be unstable and prone to overfitting spurious domain cues.}
(2) Domain-shift calibration methods \cite{wang2020transferable, park2020calibrated} use importance weighting \cite{cortes2008sample} to establish relationships between target and source samples, addressing calibration under covariate shift. Additionally, \cite{hu2024pseudo} generates pseudo-target samples to simulate supervised calibration. Despite some progress, these methods are designed for classification tasks, and such sample-level modeling strategies are not guaranteed to work well for pixel-level semantic segmentation. 
{\color{black}In particular, dense prediction requires calibrating structured, spatially correlated outputs, where domain shift can vary across regions and classes, making direct extensions of sample-level cross-domain calibration non-trivial.} 
\textit{How to achieve cross-domain calibration in semantic segmentation remains an unresolved challenge.}

These challenges motivate us to explore a novel perspective for providing appropriate supervision signals to tackle unsupervised calibration in semantic segmentation.   Our motivation stems from experimental observations of self-training-based UDA performance. When replacing argmax-based hard pseudo-labels with softmax-based soft pseudo-labels, we observe significant performance degradation. This occurs because network predictions in the target domain are poorly calibrated. In Sec. \ref{3.2}, we analyze that perfectly calibrated soft pseudo-labels should provide supervision equivalent to hard pseudo-labels.  Based on this finding, we can reframe the calibration problem in the target domain as an optimization problem for soft pseudo-labels.

To this end, we propose a cross-domain calibration framework dedicated to semantic segmentation tasks, called DA-Cal, which transforms the target domain calibration problem into soft pseudo-label optimization involving two key designs to generate appropriate calibration temperatures for semantic segmentation outputs: 
(1) To convert soft pseudo-label optimization into temperature parameter optimization, we model it as a bi-level optimization problem, inspired by meta-learning \cite{liu2018darts,finn2017model}. Specifically, we leverage bi-level optimization \cite{pham2021meta, wu2021learning} to establish the relationship between soft pseudo-labels and the UDA supervision of the segmentation network, thereby guiding the optimization direction of calibration temperature. The principle here is that if the network can produce perfectly calibrated pseudo-labels, it would also minimize the actual UDA supervision loss (i.e., the combined loss from source domain ground truth and target domain hard pseudo-labels).
(2) To effectively apply this framework to semantic segmentation tasks, we first define the temperature parameter as a Meta Temperature Network (MTN) that can adaptively output pixel-level calibration temperatures, inspired by local temperature scaling \cite{ding2021local}. Then, to effectively implement bi-level optimization, we employ complementary mixing strategies to generate different domain-mixed samples for the inner/outer optimization processes. This strategy effectively prevents MTN overfitting while mitigating domain discrepancies.
Our framework is easy to integrate into existing self-training frameworks and effectively improves the calibration of UDA segmentation networks in the target domain. Additionally, DA-Cal can bring extra  segmentation performance gains  by introducing correctly calibrated soft pseudo-labels, without additional inference overhead.

Our contributions can be summarized as follows:
(1) We identify a novel insight into UDA segmentation calibration by observing that performance degradation when using soft pseudo-labels is primarily caused by poor calibration, and analyze that perfectly calibrated soft pseudo-labels should provide supervision equivalent to hard pseudo-labels.
(2) We propose DA-Cal, a dedicated cross-domain calibration framework that transforms target domain calibration into a bi-level optimization problem, featuring a Meta Temperature Network for pixel-level calibration temperatures and complementary domain-mixing strategies to prevent overfitting while addressing the unique challenges of semantic segmentation calibration.
(3) Our method seamlessly integrates with existing self-training UDA frameworks, improving both calibration quality and segmentation performance in the target domain without additional inference overhead, demonstrating effectiveness across various UDA segmentation benchmarks.
\section{Related Work}

\subsection{Domain Adaptive Semantic Segmentation}

Unsupervised domain adaptation (UDA) for semantic segmentation addresses the costly pixel-wise annotation problem by transferring knowledge from labeled source domains to unlabeled target domains. Current approaches primarily follow two paradigms: adversarial training-based methods \cite{toldo2020unsupervised, tsai2018learning, chen2018road, ganin2015unsupervised} that align feature distributions through domain discriminators, and self-training-based methods \cite{tranheden2021dacs, hoyer2022daformer, araslanov2021self, litowards, litowards2, li2025balanced} that generate pseudo-labels within teacher-student frameworks. The latter has become dominant due to the domain-robustness of Transformers \cite{bhojanapalli2021understanding}, with researchers developing various strategies to improve pseudo-label quality, such as entropy minimization \cite{chen2019domain} and consistency regularization \cite{hoyer2023mic}. Recent advances  further expand the field: DTS \cite{huo2023focus} leverages different mixing strategies in a dual teacher-student framework, CDAC \cite{wang2023cdac} enforces consistency in attention mechanisms, and RTea \cite{zhao2023learning} incorporates structural information as additional supervision. Other innovative approaches include diffusion-based image translation \cite{peng2023diffusion}, integration of distillation with self-training \cite{shen2023diga}, utilization of domain-agnostic text embeddings \cite{mata2025copt}, and architectural innovations to learn domain-invariant representations \cite{yang2025micdrop, he2025attention}.  While these methods have significantly improved segmentation accuracy across domains, the calibration quality of UDA models remains largely unexplored.

\subsection{Uncertainty Calibration}

Calibration methods ensure model predictions align with actual correctness likelihoods, evolving from binary classification \cite{zadrozny2001obtaining, platt1999probabilistic} to multi-class settings through approaches like matrix scaling, vector scaling, and temperature scaling \cite{guo2017calibration}. Temperature scaling has become predominant due to its simplicity and effectiveness, with extensions like ATS \cite{mozafari2018attended} addressing small validation dataset challenges and BTS \cite{ji2019bin} employing bin-wise scaling with data augmentation. For semantic segmentation specifically, researchers have developed spatial-aware calibration techniques such as local temperature scaling \cite{ding2021local} to capture pixel-level heterogeneity, model ensembling with segment-level metrics \cite{mehrtash2020confidence}, and adversarial training of stochastic networks \cite{kassapis2021calibrated}. However, these supervised calibration methods rely on labeled validation data and in-distribution assumptions, making them unsuitable for unsupervised domain adaptation scenarios where labeled target data is unavailable and domain shifts are significant. To address this gap, domain-shift calibration approaches have emerged, including methods that use source validation sets as domain-generalized targets \cite{tomani2021post, salvador2021improved, zou2023adaptive}, importance weighting techniques \cite{he2008adasyn, park2020calibrated, wang2020transferable} to relate target and source samples, and multi-source domain utilization \cite{yu2022robust, gong2021confidence}. Despite this progress, most existing methods are designed for classification tasks, and their sample-level modeling strategies do not effectively translate to semantic segmentation where pixel-level calibration is required, leaving cross-domain calibration in semantic segmentation an unresolved challenge.

\noindent\textcolor{black}{\textbf{{Relation to uncertainty-aware segmentation:}}
Uncertainty-aware methods such as Bayesian deep learning \cite{gal2016dropout, kendall2015bayesian}, model ensembling \cite{lakshminarayanan2017simple}, and evidential learning \cite{sensoy2018evidential, malinin2018predictive} aim to quantify epistemic and/or aleatoric uncertainty for segmentation, which is useful for risk-aware decision making and out-of-distribution detection. These methods are conceptually related but not directly comparable to our goal: they typically require multiple stochastic forward passes (e.g., MC sampling), maintaining ensembles, or modifying the model/training objective to predict an uncertainty proxy. In contrast, DA-Cal targets \emph{probability calibration under domain shift} via a \emph{plug-and-play temperature scaling} formulation that is designed to be compatible with UDA pseudo-label training. In particular, our BI variant incurs \emph{no additional inference overhead} compared to a standard single forward pass, while still improving the reliability of confidence values used in cross-domain pseudo supervision.}

\section{Method}

\subsection{Preliminaries}
\noindent\textbf{Self-Training for UDA.}
In unsupervised domain adaptation for semantic segmentation, the network is simultaneously trained on labeled source domain data and unlabeled target domain data. Specifically, the source domain can be denoted as $D_S = {(x^S_i, y^S_i)}^{N_S}_{i=1}$, where $x^S_i \in X_S$ represents an image with $y^S_i \in Y_S$ as the corresponding pixel-wise one-hot label covering $C$ classes. Similarly, the target domain is $D_T = {(x^T_i)}^{N_T}_{i=1}$, which shares the same label space but has no access to the target labels. A self-training-based UDA pipeline consists of a supervised branch for the source domain and an unsupervised branch for the target domain. For the supervised branch, the loss $\mathcal{L}^s$ can only be calculated on the source domain to train a neural network $f_\theta$:
\begin{equation}
\label{eq:1}
\small
\mathcal{L}^s =\frac{1}{N_S} \sum_{i=1}^{N_S}\frac{1}{HW}\sum_{j=1}^{H\times W}\ell_{ce}(f_\theta(x^S_{ij}),y^S_{ij}),
\end{equation}
where $\ell_{ce}$ denotes the cross-entropy loss. The unsupervised branch introduces a teacher-student framework to generate pseudo-labels $\hat{y}^T_{ij} =\mathrm{argmax}(g_{\phi}(x^T_{ij}))$ with the teacher network $g_\phi$ for the target domain:
\begin{equation}
\label{eq.2}
\small
\mathcal{L}^u =\frac{1}{N_T} \sum_{i=1}^{N_T}\frac{1}{HW}\sum_{j=1}^{H\times W}q(p_{ij})\ell_{ce}(f_\theta(x^T_{ij}),\hat{y}^T_{ij}),
\end{equation}
where $q(p_{ij})$ is a quality estimate conditioned on confidence $p_{ij} = \max(g_{\phi}(x^T_{ij}))$ for pseudo-labels, which improves as network accuracy increases \cite{hoyer2022daformer}. After each training step, the teacher network  $g_\phi$ is updated with the exponentially moving average of the weights of $f_\theta$. The overall objective function is a combination of supervised and unsupervised losses: $\mathcal{L} = \mathcal{L}^s+\mathcal{L}^u$.

\noindent\textbf{Temperature Scaling for Calibration.} Temperature Scaling \cite{guo2017calibration} is proposed as a simple extension of Platt scaling \cite{platt1999probabilistic} for post-hoc probability calibration in multi-class classifications. Specifically, temperature scaling estimates a single scalar parameter $\mathcal{T} \in \mathbb{R}^+$, i.e., the temperature, to calibrate probabilities: $\hat{p} = \max_{c\in C} \sigma_{SM}(\mathbf{z}/\mathcal{T})^{(c)}$, where $\hat{p}$ is the calibrated probability. For semantic segmentation tasks, temperature scaling can be directly extended by estimating one global parameter $T \in \mathbb{R}^+$ for all pixels/voxels of all images:
\begin{equation}
\small
\hat{p}(x_{ij}, \mathcal{T}) = \max_{c\in C} \sigma_{SM}(f_\theta(x_{ij})/\mathcal{T})^{(c)}.
\end{equation}
The optimal value for $\mathcal{T}$ is obtained by minimizing the negative log-likelihood (NLL) with respect to a hold-out validation dataset:
\begin{align}
\small
\mathcal{T}^* = \mathop{\arg\min}\limits_{\mathcal{T}} & \left( -\sum_{i=1}^{N}\sum_{j=1}^{H\times W} \log(\sigma_{SM}(f_\theta(x_{ij})/\mathcal{T})^{(y_{ij})}) \right)  \notag  \\ & \text{s.t.}  \quad \mathcal{T} > 0.
\end{align}

However, temperature scaling in this way assumes that each image has the same distribution (i.e., the same temperature, $\mathcal{T}$, for all images), which is unrealistic, especially in domain adaptation settings where source and target distributions differ significantly.

\subsection{Motivation}
\label{3.2}
In domain adaptation classification, we analyze the relationship between hard and soft pseudo-labels.  Let $f_\theta(x)$ be a network  that outputs logits for an input $x$, and $\sigma_{SM}(f_\theta(x)/\mathcal{T})$ be the softmax probabilities after temperature scaling with parameter $\mathcal{T}$.
For a perfectly calibrated network, the predicted probability for each class should match its empirical accuracy:
\begin{equation}
\small
P(\text{correct} | \sigma_{SM}(f_\theta(x)/\mathcal{T})^{(c)} = p) = p \quad \forall p \in [0,1], \forall c \in C.
\end{equation}

This means when the network predicts class $c$ with confidence $p$, it should be correct exactly $p$ proportion of the cases. Equivalently, in terms of posterior probabilities, a perfectly calibrated network satisfies:
\begin{equation}
\small
\sigma_{SM}(f_\theta(x)/\mathcal{T})^{(c)} = P(y = c | x) \quad \forall c \in C.
\end{equation}

We now compare the losses for soft/hard pseudo-labels.

\paragraph{Soft Pseudo-Labels} When using soft pseudo-labels, the cross-entropy loss is:
\begin{equation}
\small
\mathcal{L}_{\text{soft}} = -\sum_{c \in C} \sigma_{SM}(f_\theta(x)/\mathcal{T})^{(c)} \log(\hat{p}^{(c)}).
\end{equation}

\paragraph{Hard Pseudo-Labels} For hard pseudo-labels, we select the class with the highest confidence: $\hat{y} = \arg\max_{c \in C} \sigma_{SM}(f_\theta(x)/T)^{(c)}$, yielding the loss:
\begin{equation}
\small
\mathcal{L}_{\text{hard}} = -\log(\hat{p}^{(\hat{y})}).
\end{equation}

For a perfectly calibrated network, the expected hard pseudo-label loss equals the soft pseudo-label loss:
\begin{equation}
\small
\mathbb{E}[\mathcal{L}_{\text{hard}}] = \sum_{c \in C} P(y = c | x) \cdot (-\log(\hat{p}^{(c)})) = \mathcal{L}_{\text{soft}}.
\end{equation}

% Thus, under perfect calibration, soft and hard pseudo-labels are equivalent. However, our experiments reveal severe performance degradation when using only soft pseudo-labels in UDA. This degradation stems from miscalibration of the soft pseudo-labels. Therefore, we can reformulate the calibration problem as an optimization task aimed at improving soft pseudo-label quality.

Thus, under perfect calibration, soft and hard pseudo-labels are equivalent \emph{in expectation}.
{\color{black}{\color{black}Importantly, this provides a \emph{sufficient} condition: perfect calibration implies soft/hard equivalence in expectation, and therefore soft labels are not inherently weaker in terms of expected supervision strength under this condition.}}
{\color{black}Conversely, when a substantial performance gap is observed between soft-only and hard pseudo-label training, we do not invoke the (generally false) converse statement. Instead, we use the logically valid contrapositive of the above implication: if soft-only training is \emph{not} equivalent to hard pseudo-label training in practice (e.g., exhibits a clear performance drop), then the sufficient condition must be violated, indicating that the soft pseudo-labels are \emph{not} perfectly calibrated.}
However, our experiments reveal severe performance degradation when using only soft pseudo-labels in UDA. {\color{black}By the contrapositive argument above, this observation provides evidence that the soft pseudo-label probabilities are not perfectly calibrated, and the dominant gap is attributable to miscalibration rather than inherently weaker expected supervision.} Therefore, we can reformulate the calibration problem as an optimization task aimed at improving soft pseudo-label quality.
{\color{black}Meanwhile, even after calibration, soft and hard labels may still exhibit slightly different optimization dynamics (i.e., error-propagation behaviors), which can lead to a small residual gap; this is empirically discussed in Tab. \ref{table4}.}

% {\color{black}
% \textbf{Discussion: Why calibration matters for stable self-training:}
% In UDA segmentation, self-training typically uses model predictions as soft supervision. When predictions are miscalibrated, the induced soft labels allocate non-trivial probability mass to incorrect classes. This does not only harm the quality of pseudo-labels, but also biases the optimization signal: the gradient of the cross-entropy with respect to logits is \( \nabla_z \ell = p - y \). If the soft target \(y\) (or its hard counterpart obtained by thresholding) is derived from miscalibrated probabilities, then \(y\) deviates from the ideal supervision, and the resulting gradient direction deviates from the desired one, which makes optimization less stable and amplifies confirmation bias.
% }

{\color{black}
\textbf{Discussion: miscalibration induces unstable optimization in self-training:}
In UDA segmentation, self-training pipelines typically rely on \emph{hard} pseudo-labels (e.g., \(\hat{y}^{T}_{ij}=\arg\max(g_{\phi}(x^{T}_{ij}))\) in Eq.~\eqref{eq.2}) with confidence-based filtering \(q(p_{ij})\) to mitigate noise. We discuss \emph{soft} pseudo-labels here only as a theoretical lens to understand when they should be as effective as hard labels (under perfect calibration) and why they can fail in practice (under miscalibration).
Concretely, under miscalibration, the soft pseudo-label distribution assigns non-trivial probability mass to incorrect classes. This corrupts the supervision signal and biases the optimization direction. For a pixel-wise cross-entropy loss with soft targets \(y\) and predicted probabilities \(p\), the gradient w.r.t. the logits is \( \nabla_{z}\ell_{ce} = p - y \). Hence, when \(y\) is derived from miscalibrated probabilities, it deviates from the ideal posterior \(P(y|x)\), and the resulting gradient direction deviates accordingly, which can amplify confirmation bias and make self-training less stable. This motivates treating calibration as an optimization problem whose goal is to improve the quality of pseudo-label supervision.}

\subsection{The DA-Cal Framework}
% To effectively address the pseudo-label calibration problem in domain adaptation, we propose a bi-level optimization framework called DA-cal (shown in Fig. \ref{fig2}). This framework simultaneously optimizes the segmentation network and the calibration network through meta-learning. The method is inspired by bi-level optimization, consisting of inner optimization and outer optimization steps. The inner optimization focuses on learning optimal temperature calibration parameters, while the outer optimization updates the segmentation network using the calibrated pseudo-labels. 

\begin{figure}
    \centering
    \includegraphics[width=1\linewidth]{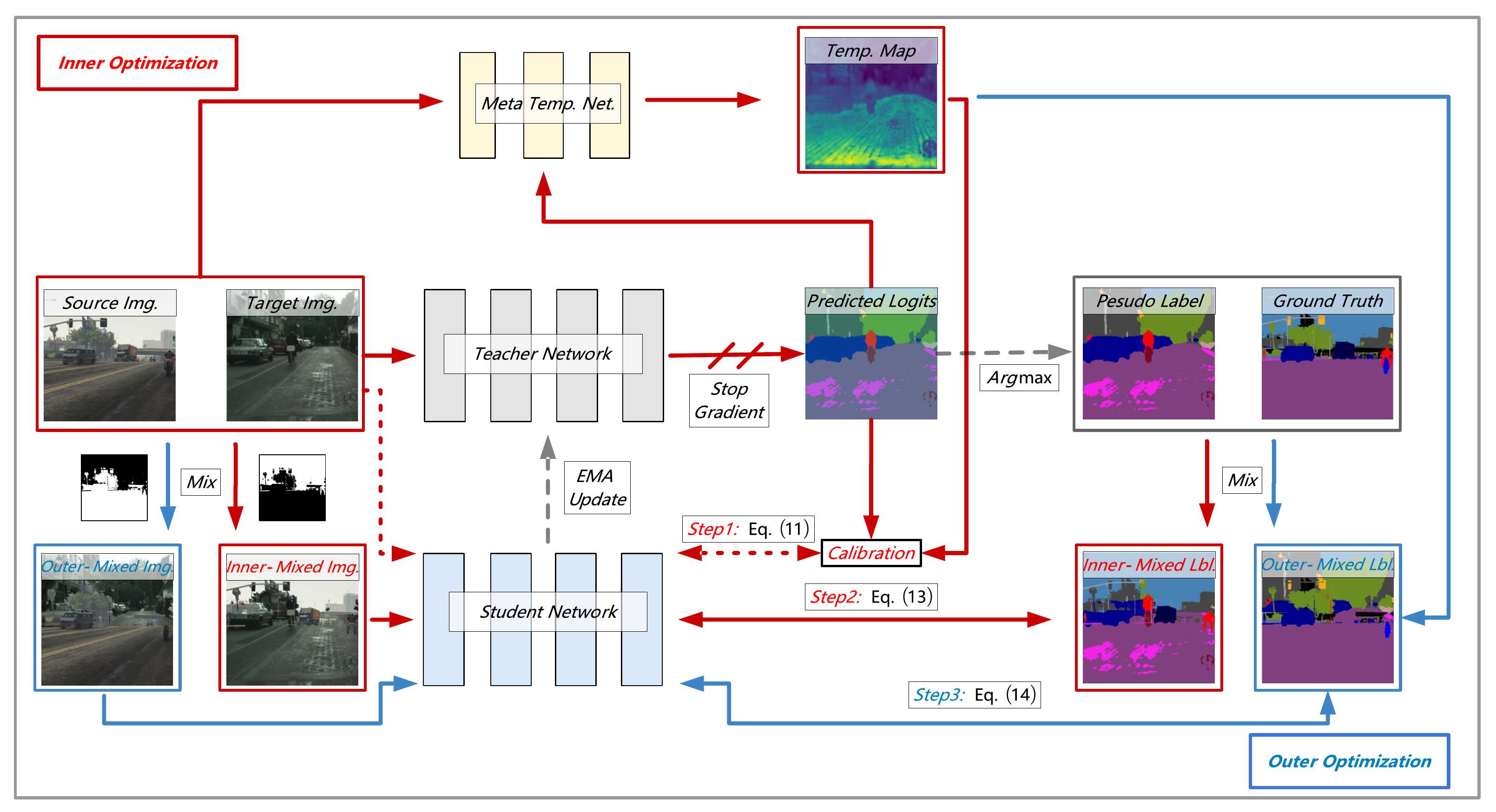}
    \caption{ \textbf{Pipeline illustration of DA-Cal}. Our framework consists of bi-level optimization: inner optimization (Steps 1-2) for calibrating predictions with Meta Temperature Network, and outer optimization (Step 3) for training the segmentation network with both calibrated soft and hard pseudo-labels. The supervised loss on source domain is omitted for clarity. }
    \label{fig2}
 % \vspace{-1em}
\end{figure}

To address pseudo-label miscalibration in domain adaptation, we propose DA-Cal, a bi-level optimization framework (illustrated in Fig. \ref{fig2}). DA-Cal simultaneously optimizes the segmentation network and the calibration network inspired by  meta-learning. It consists of:
\begin{itemize}
    \item \textit{Inner optimization}, which learns optimal temperature calibration parameters.
    \item \textit{Outer optimization}, which updates the segmentation network using calibrated pseudo-labels.
\end{itemize}

\noindent\textbf{Meta Temperature Network (MTN).} 
We introduce a  Meta Temperature Network (MTN) to predict pixel-wise temperature values  for calibration. The network simply consists of stacked Conv-BatchNorm-ReLU blocks for efficient processing. The MTN takes two inputs: the original image and the logits output by the segmentation network, which are concatenated along the channel dimension. The MTN outputs a temperature map of the same spatial size, assigning a dedicated temperature $\mathcal{T}_{ij}$ to each pixel: 
\begin{equation}
\small
{\color{black}p_{ij,\text{cal}} }= \sigma_{\text{SM}}(f_\theta(x_{ij})/\mathcal{T}_{ij}).
\end{equation}
This fine-grained calibration adapts to  varying uncertainty levels across different regions, resulting in more accurate soft pseudo-labels.
Notably, MTN is designed as a domain-shared module, leveraging source domain calibration knowledge to guide target domain calibration.

\noindent\textbf{Inner Optimization.} 
The inner optimization consists of two key steps:

\textit{\textbf{Step 1: Learning from Temperature-Calibrated Pseudo-Labels.}} We first create a temporary copy $\theta'$ of the student network parameters $\theta$. Then, we calibrate both source and target domain predictions using  MTN. For both domains, we utilize calibrated soft pseudo-labels for training. 
A one-step gradient update of $\theta'$ is performed based on these calibrated losses:
\begin{equation}
\small
\theta'^{[k+1]} = \mathop{\arg\min}\limits_{\theta'} \mathcal{L}_{\text{cal}}(S, T; \theta'^{[k]}, \psi^{[k]}),
\end{equation}
which can be approximated as:
\begin{equation}
\label{eq:11}
\small
\theta'^{[k+1]} = \theta'^{[k]} - \alpha \nabla_{\theta'^{[k]}}\mathcal{L}_{\text{cal}}(S, T; \theta'^{[k]}, \psi^{[k]}),
\end{equation}
where $\psi$ represents the MTN parameters and $\alpha$ is the inner-loop learning rate. Here, $\mathcal{L}_{\text{cal}}$ is the calibrated cross-entropy loss based on soft pseudo-labels.

\begin{figure*}[t]
\centering
\resizebox{1\linewidth}{!}{
\begin{minipage}{\linewidth}
\begin{algorithm}[H]
\caption{Pseudo algorithms of DA-Cal.}
\label{alg:1}
\begin{algorithmic}[1]
\State \textbf{Inputs:}
Source Domain $D_S = {(x^S_i, y^S_i)}^{N_S}_{i=1}$, Target Domain $D_T = {(x^T_i)}^{N_T}_{i=1}$
\State \textbf{Define:} Student Network $f_\theta$, Teacher Network $g_\phi$, Meta Temperature Network MTN $h_\psi$, Inner-loop Learning Rate $\alpha$, MTN Learning Rate $\beta$,  Outer-loop Learning Rate $\delta$, EMA Coefficient $\gamma$
\State \textbf{Output:} Student Network $f_\theta$
\For{each batch of $(x^S_i, y^S_i)$, $x^T_i$ in $D_s$, $D_t$}
\State \textbf{\textit{\# Inner Optimization:}}
\State Create temporary copy $\theta'^{[k]}  \leftarrow \theta^{[k]} $ \Comment{\textit{ Step 1}}
\State Generate logits $z^S_i = g_{\phi^{[k]} }(x^S_i)$, $z^T_i = g_{\phi^{[k]}}(x^T_i)$
\State Predict temperatures $\mathcal{T}^S_i = h_{\psi^{[k]}} ([x^S_i, z^S_i])$, $\mathcal{T}^T_i = h_{\psi^{[k]}} ([x^T_i, z^T_i])$
\State Calibrate predictions $p^S_{i,cal} = \sigma_{SM}(z^S_i/\mathcal{T}^S_i)$, $p^T_{i,cal} = \sigma_{SM}(z^T_i/\mathcal{T}^T_i)$

\State Update $\theta'^{[k]}$ using gradient descent: $\theta'^{[k+1]} \leftarrow \theta'^{[k]}  - \alpha \nabla_{\theta'^{[k]} }\mathcal{L}_{\text{cal}}(S, T; \theta'^{[k]} , \psi^{[k]} )$ by Eq. \ref{eq:11}

\State Generate  complementary mixed dataset ( [\textit{inner/outer-mixed image}]) using ClassMix \Comment{Step 2}

\State Update MTN parameters on [\textit{inner-mixed image}]: $\psi^{[k+1]} \leftarrow \psi^{[k]}  - \beta \nabla_{\psi^{[k]} } \mathcal{L}_{\text{mix}}(\theta'^{[k+1]} )$ by Eq. \ref{eq:13}
\State \textbf{\textit{\# Outer Optimization:}}
\State Calculate supervised loss $\mathcal{L}^S$ using ground truth labels $y^S_i$ by Eq. \ref{eq:1} 
\State Generate target pseudo-labels $\hat{y}^T_i = \text{argmax}(g_\phi(x^T_i))$
\State Predict calibrated soft pseudo-labels $p^T_{i,cal} = \sigma_{SM}(f_\theta(x^T_i)/h_{\psi^{[k+1]} }([x^T_i, f_\theta(x^T_i)]))$  \Comment{Step 3}
\State Calculate unsupervised loss $\hat{\mathcal{L}}^u$ combining hard and soft pseudo-labels  by Eq. \ref{eq:14}  

\State Update student network on both source image and [\textit{outer-mixed image}]: $\theta^{[k+1]} \leftarrow \theta^{[k]} - \delta\nabla_{\theta^{[k]}}(\mathcal{L}^S + \hat{\mathcal{L}}^u)$

\State $\phi^{[k+1]} \leftarrow \gamma \phi^{[k]} + (1-\gamma)\theta^{[k+1]}$ \Comment{\textit{EMA Update}}
\EndFor

\end{algorithmic}
\end{algorithm}
\end{minipage}}
 \vspace{-1em}
\end{figure*}

\textit{\textbf{Step 2:  Updating MTN via Meta-Learning.}} To evaluate the effectiveness of calibration, we test the updated network  parameters $\theta'^{[k+1]}$ on a specially constructed mixed dataset. This mixed dataset is created using ClassMix \cite{olsson2021classmix} technique, generating samples that are complementary to those used in the outer optimization to prevent self-reinforcement and overfitting. The performance on this mixed dataset reflects the network's domain adaptation capability. We compute the loss of $\theta'^{[k+1]}$ on this mixed dataset and update the MTN parameters $\psi$ through the meta-learning process:
\begin{equation}
\label{eq:12}
\small
\psi^{[k+1]} = \mathop{\arg\min}\limits_{\psi^{[k]}} \mathcal{L}_{\text{mix}}(\theta'^{[k+1]}),
\end{equation}
which is implemented as:
\begin{equation}
\label{eq:13}
\small
\psi^{[k+1]} = \psi^{[k]} - \beta \nabla_{\psi^{[k]}} \mathcal{L}_{\text{mix}}(\theta'^{[k]} - \alpha \nabla_{\theta'^{[k]}}\mathcal{L}_{\text{cal}}(\theta'^{[k]}, \psi^{[k]} ),
\end{equation}
where $\beta$ is the learning rate for MTN. The $\mathcal{L}_{\text{mix}}$ is the cross-entropy loss computed on the mixed dataset, which combines source domain ground truth labels and target domain hard pseudo-labels. This mixed supervision provides a  learning signal that helps evaluate how well the calibration improves the network's domain adaptation performance.

Since the MTN is domain-shared, calibration knowledge learned from the source domain (where ground truth is available) can gradually guide the calibration learning in the target domain. 

\noindent\textbf{Outer Optimization.} 
After completing the inner optimization, we proceed with outer optimization to update the segmentation network parameters:

\textit{\textbf{Step 3: Updating the Segmentation Network.}} The network is trained using  the following total loss function: $\mathcal{L}_{\text{total}} = \mathcal{L}^{s} + \hat{\mathcal{L}}^{u},$
where $\mathcal{L}^{s}$ is the supervised loss on the source domain computed with ground truth labels, and $\hat{\mathcal{L}}^{u}$ is the unsupervised loss on the target domain. Unlike Eq. \ref{eq.2},  $\hat{\mathcal{L}}^{u}$ combines both hard and soft pseudo-label losses:
\begin{equation}
\begin{aligned}
\label{eq:14}
\small
\hat{\mathcal{L}}^u  =\frac{1}{N_T} \sum_{i=1}^{N_T}\frac{1}{HW}\sum_{j=1}^{H\times W}q(p_{ij})  \left(\ell_{ce}(f_\theta(x^T_{ij}),\hat{y}^T_{ij})  \right. \\
 + \ell_{ce}(f_\theta(x^T_{ij}),p^{T}_{ij,\text{cal}}) \left. \right ).
\end{aligned}
\end{equation}

% The calibrated soft pseudo-labels provide complementary supervision by preserving prediction uncertainty that hard labels discard. This framework not only achieves effective calibration but also brings additional segmentation performance gains by reducing confirmation bias in self-training. The  algorithmic description  is  detailed in Alg. \ref{alg:1}.

The calibrated soft pseudo-labels \(p^{T}_{ij,\text{cal}}\) in Eq.~\eqref{eq:14} provide complementary supervision by preserving uncertainty that hard labels discard{\color{black}, while still being anchored by hard pseudo-label learning. More importantly, DA-Cal is \emph{not} designed to directly minimize a calibration metric such as ECE. Instead, it minimizes an \emph{outcome-driven surrogate objective} defined by the bi-level procedure: the MTN \(h_\psi\) is updated through the meta-loss \(\mathcal{L}_{\text{mix}}\) on complementary mixed data (Eq.~\eqref{eq:13}), which evaluates whether a calibration update actually improves UDA supervision after cross-domain transfer. As a result, MTN updates that merely reshape confidence without improving downstream self-training are discouraged, whereas updates that lead to better adaptation performance are reinforced. This is why DA-Cal can improve both calibration and segmentation accuracy in self-training-based UDA. See Appendix~\ref{sec:bilevel_formulation} for the  theoretical formulation of bi-level optimization, and}
the  algorithmic description  is  detailed in Alg. \ref{alg:1}.

{\color{black}
Although the proposed MTN takes both the input image and the corresponding logits as input---which could, in principle, allow it to exploit domain-specific appearance patterns or confidence statistics---our design explicitly discourages such shortcuts. First, our \emph{complementary mixing} constructs samples that contain regions from both the source and target domains, thereby weakening the predictiveness of purely domain-indicative signals and encouraging transformations that remain valid across domains. Second, MTN is \emph{domain-shared} rather than employing domain-specific heads, which limits the capacity to memorize domain-dependent behaviors. Third, MTN is optimized with an \emph{outcome-driven meta-objective}: updates are retained only if they improve calibration after cross-domain transfer, making reliance on domain cues unhelpful unless it consistently yields better target-domain calibration.}

\subsection{Practical Implementation}
{\color{black}
\subsubsection{Optimization stability and computational overhead}
\label{sec:overhead_stability}
DA-Cal adopts a bi-level update with an inner self-training step and an outer meta-objective on complementary mixed data. The inner update produces
$\theta'(\psi)=\theta-\alpha\nabla_{\theta}\mathcal{L}_{\text{inner}}(\theta,\psi),$
and MTN is updated by the meta-gradient \(\nabla_{\psi}\mathcal{L}_{\text{mix}}(\theta'(\psi))\). This gradient may include higher-order terms due to differentiating through \(\theta'(\psi)\). To ensure scalability, we restrict the differentiable inner-loop update to the \emph{segmentation head} only and keep the backbone frozen, so the higher-order computation is confined to a small parameter set (Appendix~\ref{sec:C}). In addition, MTN is lightweight (Appendix~~\ref{sec:E}), so the extra compute introduced by the calibration module itself is modest.

In practice, the main overhead comes from an additional forward/backward associated with the one-step head update and the meta-loss on mixed samples. We report the measured GPU memory usage and training-time overhead in  Tab.~\ref{tab:efficiency}, showing that DA-Cal adds only a limited computational overhead while maintaining stable optimization.}

\subsubsection{\color{black} Implementation strategies}
We offer two implementation strategies for our calibration approach: \textbf{DA-Cal-PH} and \textbf{DA-Cal-BI}, corresponding to post-hoc strategy and built-in strategy respectively.

\noindent\textbf{DA-Cal-PH (Post-Hoc).}  
In this implementation,  MTN is used to calibrate network outputs during inference. Notably, MTN does not affect network accuracy, allowing it to be omitted when calibration is unnecessary.  This provides flexibility for deployment in different application contexts.

\noindent\textbf{DA-Cal-BI (Built-In).}  
This implementation integrates calibration directly into the loss function by 
incorporating the MTN-predicted temperature $\mathcal{T}$ into the cross-entropy loss:  $\ell_{ce}(x, y) = -\log\left({\sigma_{SM}(f_\theta(x)*\mathcal{T})}^{(y)}\right)/\mathcal{T}$. 
Here, the network effectively learns a scoring function of the form $g(x) = f(x)*\mathcal{T}$. Consequently, since $\text{argmax}f(x) = g(x)/\mathcal{T}$,  this formulation is analogous to post-hoc adjustment, enabling the network to produce calibrated outputs directly.
The key advantage of DA-Cal-BI is that it produces a pre-calibrated network, eliminating the need for soft pseudo-labels in Eq. \ref{eq:14} and avoiding any additional inference overhead.

\section{Experiments}

\begin{table*}[ht]

\small
    \centering
    \caption{Calibration Metrics of UDA on Autonomous Driving Scenarios. Note that these metrics on SYN.  are calculated over 16 classes.} \label{table1}
       \setlength{\tabcolsep}{3.5pt} 
    \begin{tabular}{l|ccc|ccc|ccc|ccc}
    \hline
    \multirow{2}{*}{\textbf{Method}} 
    & \multicolumn{3}{c|}{\textbf{DACS \cite{tranheden2021dacs}}} 
    & \multicolumn{3}{c|}{\textbf{DAFormer \cite{hoyer2022daformer}}} 
    & \multicolumn{3}{c|}{\textbf{MIC \cite{hoyer2023mic}}}
    & \multicolumn{3}{c}{\textbf{\color{black}DiDA \cite{litowards}}} \\
    \cline{2-13}
    & ECE ($\downarrow$) & NLL ($\downarrow$) & BS ($\downarrow$)
    & ECE ($\downarrow$) & NLL ($\downarrow$) & BS ($\downarrow$)
    & ECE ($\downarrow$) & NLL ($\downarrow$) & BS ($\downarrow$)
    & {\color{black}ECE ($\downarrow$)} & {\color{black}NLL ($\downarrow$)} & {\color{black}BS ($\downarrow$)}\\
 
    \hline
    \hline
    \multicolumn{13}{c}{\textbf{GTAv$\rightarrow$Cityscapes}} \\
    \hline
\rowcolor{black!5}    NoAdapt & 0.2937& 2.8097&0.8213 & 0.2098&1.7237 & 0.6262& 0.1806&1.5172 &  0.6209 &    {\color{black}0.1806} & {\color{black}1.5172} & {\color{black}0.6209}\\
    \hline
    No Calib. & 0.2289&1.7445 & 0.5701& 0.1255 & 0.9084&0.3456 & 0.1042 & 0.7659 & 0.3034 & {\color{black}0.0907} & {\color{black}0.7089} & {\color{black}0.2864} \\

        Ensemble &0.1703 & 1.5054& 0.5400& 0.0914&  0.7325&  0.3212& 0.0654&0.6971&0.2828 & {\color{black}0.0609} & {\color{black}0.6748} & {\color{black}0.2733}\\

            {\color{black}MC Dropout} &
{\color{black}0.1896} & {\color{black}1.6037} & {\color{black}0.5579} &
{\color{black}0.1038} & {\color{black}0.7814} & {\color{black}0.3341} &
{\color{black}0.0712} & {\color{black}0.7129} & {\color{black}0.2924}  &{\color{black}-} & {\color{black}-} & {\color{black}-} \\

    TempScal-src & 0.1877&  1.4393&0.5375 &0.1030 &0.8019 &0.3325 &  0.0785& 0.6682&0.2909   & {\color{black}0.0728} & {\color{black}0.6486} & {\color{black}0.2811} \\

{\color{black}TempScal-src-LTS} & {\color{black}0.1778}& {\color{black}1.3896}&{\color{black}0.5287} & {\color{black}0.0964} &{\color{black}0.7721}&{\color{black}0.3269} & {\color{black}0.0742}&{\color{black}0.6489}&{\color{black}0.2860} &{\color{black}-} & {\color{black}-} & {\color{black}-} \\
 
    PseudoCal &0.1580 &1.2893 &0.5141 & 0.0818 & 0.7431 & 0.3242 & 0.0595&  0.6242&0.2826 &{\color{black}0.0538} & {\color{black}0.6067} & {\color{black}0.2695} \\

         {\color{black}PseudoCal-LTS} & {\color{black}0.1491}&{\color{black}1.2448}&{\color{black}0.5042} & {\color{black}0.0762}&{\color{black}0.7187}&{\color{black}0.3188} & {\color{black}0.0564}&{\color{black}0.6038}&{\color{black}0.2786}&{\color{black}-} & {\color{black}-} & {\color{black}-} \\
         
 \rowcolor{blue!5}   \textbf{DA-Cal (ours)} & \textbf{0.0965}& \textbf{0.8327}& \textbf{0.3279}& \textbf{0.0578} &\textbf{0.6332} & \textbf{0.2550}&  \textbf{0.0459}&\textbf{0.5443}&\textbf{0.2116}   & {\color{black}\textbf{0.0408}} & {\color{black}\textbf{0.5162}} & {\color{black}\textbf{0.2019}} \\
    \hline
\rowcolor{yellow!5}    Oracle &0.0932 &0.7626 &0.3060 & 0.0632& 0.5760&0.2314&0.0592 &0.5390 & 0.2044    & {\color{black}0.0592} & {\color{black}0.5390} & {\color{black}0.2044}\\
    \hline
    \hline
    \multicolumn{13}{c}{\textbf{SYNTHIA$\rightarrow$Cityscapes}} \\
    \hline
\rowcolor{black!5}     NoAdapt &0.3643 &3.3656&0.8939 & 0.2705 &  2.4720&  0.7061&0.2359 &1.9054 &0.5787    & {\color{black}0.2359} & {\color{black}1.9054} & {\color{black}0.5787} \\
    \hline
    No Calib. & 0.2730& 2.3052 & 0.6516 & 0.1868 &   1.6878&   0.4406 &0.1668 & 1.2507&0.4763  & {\color{black}0.1548} & {\color{black}1.2059} & {\color{black}0.4627}\\
        Ensemble &0.2030 &1.8273 &0.5576&0.1449 &1.3538 & 0.3935&0.1056 &1.1054 &0.3716   & {\color{black}0.0989} & {\color{black}1.0728} & {\color{black}0.3624}\\
    TempScal-src & 0.2388 &  1.9255&  0.6236& 0.1658& 1.4204& 0.4234 & 0.1356&1.1332 & 0.4217 & {\color{black}0.1287} & {\color{black}1.1046} & {\color{black}0.4021} \\

 {\color{black}TempScal-src-LTS} & {\color{black}0.2270}&{\color{black}1.8641}&{\color{black}0.6121} & {\color{black}0.1587}&{\color{black}1.3746}&{\color{black}0.4143} & {\color{black}0.1298}&{\color{black}1.1019}&{\color{black}0.4135}&{\color{black}-} & {\color{black}-} & {\color{black}-} \\

    PseudoCal &  0.2094& 1.6939& 0.6008&0.1447 & 1.2366& 0.4108& 0.1229&1.0870 &0.4075 & {\color{black}0.1165} & {\color{black}1.0504} & {\color{black}0.3929} \\

  {\color{black}PseudoCal-LTS} & {\color{black}0.1986}&{\color{black}1.6397}&{\color{black}0.5889} & {\color{black}0.1389}&{\color{black}1.2028}&{\color{black}0.4025} & {\color{black}0.1187}&{\color{black}1.0554}&{\color{black}0.3988} &{\color{black}-} & {\color{black}-} & {\color{black}-} \\
    
  \rowcolor{blue!5}    \textbf{DA-Cal (ours)} &\textbf{0.1530} &\textbf{1.2377} & \textbf{0.4652}& \textbf{0.0950}&\textbf{0.8155} & \textbf{0.3213} &\textbf{0.0822} & \textbf{0.8038} &\textbf{0.3256}      &{\color{black}\textbf{0.0766}} & {\color{black}\textbf{0.7749}} & {\color{black}\textbf{0.3148}} \\
    \hline
 \rowcolor{yellow!5}    Oracle  &0.0932 &0.7626 &0.3060 &0.0632& 0.5760&0.2314 &0.0592 &0.5390 & 0.2044     & {\color{black}0.0592} & {\color{black}0.5390} & {\color{black}0.2044} \\
    \hline
    \hline
    \multicolumn{13}{c}{\textbf{Cityscapes$\rightarrow$ACDC}} \\
    \hline
 \rowcolor{black!5}   NoAdapt & 0.4431 &4.7043 &0.9792 & 0.3265&  3.3191&  0.7355& 0.2584& 2.7643& 0.6267  &{\color{black}0.2584} & {\color{black}2.7643} & {\color{black}0.6267}\\
    \hline
    No Calib. &0.3245 &2.9316 & 0.7893 &0.2282 & 1.9561& 0.5350& 0.1886&1.4702 & 0.4852 & {\color{black}0.1761} & {\color{black}1.4212} & {\color{black}0.4746}\\
       Ensemble & 0.2565&2.3739 & 0.7326&0.1773 & 1.5063&0.4719 & 0.1456& 1.2139& 0.4234 & {\color{black}0.1412} & {\color{black}1.1883} & {\color{black}0.4171}\\
    TempScal-src &0.2807 &2.4759 &0.7529 & 0.2015&1.6227& 0.5106&0.1567 &1.3380 &0.4569 & {\color{black}0.1499} & {\color{black}1.2927} & {\color{black}0.4419} \\

 {\color{black}TempScal-src-LTS} & {\color{black}0.2656}&{\color{black}2.3864}&{\color{black}0.7368} & {\color{black}0.1929}&{\color{black}1.5613}&{\color{black}0.4988} & {\color{black}0.1489}&{\color{black}1.2915}&{\color{black}0.4476} &{\color{black}-} & {\color{black}-} & {\color{black}-} \\
  
    PseudoCal & 0.2375&2.1926 &0.7221 &  0.1763& 1.4173 & 0.4938 &0.1463 &1.2890 &0.4416 & {\color{black}0.1391} & {\color{black}1.2449} & {\color{black}0.4310} \\

 {\color{black}PseudoCal-LTS} & {\color{black}0.2239}&{\color{black}2.1218}&{\color{black}0.7076} & {\color{black}0.1678}&{\color{black}1.3682}&{\color{black}0.4821} & {\color{black}0.1396}&{\color{black}1.2517}&{\color{black}0.4325} &{\color{black}-} & {\color{black}-} & {\color{black}-} \\
      
  \rowcolor{blue!5}    \textbf{DA-Cal (ours)} &\textbf{0.1874} & \textbf{1.7141}& \textbf{0.5953}& \textbf{0.1216}& \textbf{1.0034}& \textbf{0.3732} &\textbf{0.1056} &\textbf{0.0913} & \textbf{0.3934} & {\color{black}\textbf{0.0986}} & {\color{black}\textbf{0.9341}} & {\color{black}\textbf{0.3618}} \\

    \hline
\rowcolor{yellow!5}     Oracle &0.1842 &1.5083 &0.4738 & 0.1255 & 0.9948 & 0.3423& 0.0884&0.7644 & 0.3167   & {\color{black}0.0884} & {\color{black}0.7644} & {\color{black}0.3167}\\
    \hline
    \end{tabular}
      \vspace{-1em}
\end{table*}

\subsection{Experimental Setup}
\noindent \textbf{Datasets.} 
We evaluate our calibration approach on multiple benchmarks across two application domains where accurate uncertainty estimation is critical: autonomous driving and biomedical imaging.
\textbf{For autonomous driving}, we consider both synthetic-to-real and clear-to-adverse weather domain adaptation scenarios. As synthetic datasets, we use GTAv \cite{richter2016playing}, containing 24,966 images, and SYNTHIA \cite{ros2016synthia}, with 9,400 images. For real-world domains, we employ Cityscapes \cite{cordts2016cityscapes}, which provides 2,975 training images and 500 validation images under clear weather conditions, and ACDC \cite{sakaridis2021acdc}, which includes 1,600 training, 406 validation, and 2,000 test images captured under adverse weather conditions (fog, night, rain, and snow).
\textbf{For biomedical imaging}, we evaluate on three widely used electron microscopy (EM) datasets exhibiting significant domain shifts. The VNC III dataset \cite{gerhard2013segmented} contains 20 sections (1024 $\times$ 1024) of Drosophila brain tissue imaged using serial-section transmission electron microscopy (ssTEM). The Lucchi dataset \cite{lucchi2013learning} consists of two volumes (165 $\times$ 1024 $\times$ 768) from mouse brain, acquired via focused ion beam scanning electron microscopy (FIB-SEM). The MitoEM dataset \cite{wei2020mitoem} comprises two large-scale volumes (1000 $\times$ 4096 $\times$ 4096) from rat and human samples, imaged with multi-beam scanning electron microscopy (mbSEM).

\noindent \textbf{Evaluation Metrics.}  For calibration quality, we consider three metrics: Expected Calibration Error (ECE) \cite{guo2017calibration}, Negative Log-Likelihood (NLL) \cite{goodfellow2016deep}, and Brier Score (BS) \cite{brier1950verification}. These metrics assess different aspects of prediction reliability and confidence alignment with actual accuracy. Note that, different from classification task, semantic segmentation has a significant class-pixel imbalance issue; therefore, we compute the above metrics from a class-pixel balanced perspective. For segmentation performance, we use mean Intersection over Union (mIoU) to evaluate the accuracy of semantic class predictions. Detailed descriptions of these metrics are provided in Appendix
\ref{sec:A}.

\noindent \textbf{UDA Methods.} For autonomous driving benchmarks, i.e., GTAv$\rightarrow$Cityscapes, SYNTHIA$\rightarrow$Cityscapes, and Cityscapes$\rightarrow$ACDC, we adopt four baselines of increasing strength: DACS \cite{tranheden2021dacs}, DAFormer \cite{hoyer2022daformer},  MIC \cite{hoyer2023mic}, {\color{black}and DiDA \cite{litowards}.} DACS  uses the DeepLabV2 \cite{chen2017deeplab} network based on ResNet-101 \cite{he2016deep}, while the latter {\color{black}three} employ MiT-B5 \cite{xie2021segformer} as the backbone. For biomedical imaging benchmarks, including VNC III$\rightarrow$Lucchi (Subset1) and MitoEM-R$\rightarrow$MitoEM-H, we adopt two baselines, DA-ISC \cite{huang2022domain} and CAFA \cite{yin2023class}, following the U-Net \cite{ronneberger2015u} structure.

% \noindent \textbf{Calibration Baselines.} For a comprehensive comparison, we consider 4 calibration baselines, including the no calibration baseline (No Calib.), unsupervised calibration method (Ensemble \cite{mehrtash2020confidence, lakshminarayanan2017simple}), source-domain calibration (TempScal-src \cite{guo2017calibration}), and cross-domain calibration (PsudoCal \cite{hu2024pseudo}) We also report metrics separately trained on source and target domains (NoAdapt/Oracle). Details of these implementation are povided in Appendix  \ref{sec:B}.

\noindent \textbf{Calibration Baselines.} For a comprehensive comparison, we consider {\color{black}6} calibration baselines, including the no calibration baseline (No Calib.), unsupervised calibration method (Ensemble~\cite{mehrtash2020confidence, lakshminarayanan2017simple}{\color{black}, and MC Dropout \cite{gal2016dropout}}), source-domain calibration (TempScal-src~\cite{guo2017calibration}{\color{black}, and its segmentation-specific extension Local Temperature Scaling (TempScal-src-LTS~\cite{ding2021local})}), and cross-domain calibration (PseudoCal~\cite{hu2024pseudo}{\color{black}, as well as its LTS variant (PseudoCal-LTS)}). We also report metrics separately trained on source and target domains (NoAdapt/Oracle). Details of these implementations are provided in Appendix~\ref{sec:B}.

\noindent \textbf{Implementation Details.} We train all UDA methods using their official code. For Inner Optimization and MTN network, we adopt the SGD optimizer with a fixed learning rate of 0.01. For  biomedical imaging benchmarks, we use CutMix \cite{yun2019cutmix} instead of ClassMix \cite{olsson2021classmix} since these are binary classification tasks. More training and implementation details are provided in Appendix \ref{sec:C}.

\begin{table}[t]

\small
    \centering
    \caption{Calibration Metrics of UDA on   Biomedical Imaging  Scenarios, which  are calculated only on the foreground.} \label{table2}
    \setlength{\tabcolsep}{2pt}
    \begin{tabular}{l|ccc|ccc}
     \toprule
        \multirow{2}{*}{\textbf{Method}} & \multicolumn{3}{c|}{\textbf{DA-ISC \cite{huang2022domain}}} & \multicolumn{3}{c}{\textbf{CAFA \cite{yin2023class}}} \\ 
        \cline{2-7}
        &ECE &NLL &BS &ECE &NLL &BS 
         \\
             \hline 
             \hline 
      \multicolumn{7}{c}{\textbf{VNC III $\rightarrow$ Lucchi (Subset1)}}\\
    \hline
 \rowcolor{black!5}    NoAdapt & 0.3182& 3.5611& 0.6522& 0.3182& 3.5611& 0.6522 \\
    \hline
    No Calib. &0.2678& 2.1510 &0.5501& 0.2167&2.0625 & 0.4445\\
       Ensemble  &0.1530& 1.0610& 0.4662 &0.0869 &0.7738 &0.3544 \\
    TempScal-src   &0.2011 & 0.8115& 0.4726&0.1744 & 0.7489&0.3968\\

    PseudoCal &0.1219&0.6202& 0.4158 & 0.1086& 0.5499& 0.3531 \\
  \rowcolor{blue!5}   \textbf{DA-Cal (ours)}& \textbf{0.0585}& \textbf{0.5770}& \textbf{0.3919} &\textbf{0.0520} & \textbf{0.5057}& \textbf{0.3321} \\
    \hline
\rowcolor{yellow!5}     Oracle & 0.0471& 0.5023& 0.3281& 0.0471& 0.5023& 0.3281\\

             \hline 
             \hline 
      \multicolumn{7}{c}{\textbf{MitoEM-R$\rightarrow$ MitoEM-H}}\\
        \hline
 \rowcolor{black!5}      NoAdapt &0.2072 & 0.8347&0.6202&0.2072 & 0.8347&0.6202  \\
    \hline
    No Calib. &0.1983&1.5984&0.4089 &0.1404& 1.3978 &0.2897\\
        Ensemble &0.0948 & 0.5371&0.3217 &0.0710 & 0.5435&0.2557 \\
    TempScal-src &0.1385 &0.6154&0.3531 &0.1097&0.5101 &0.2616  \\

    PseudoCal & 0.0702 & 0.5049&0.3385 & 0.0689&0.4116 & 0.2475 \\
  \rowcolor{blue!5}    \textbf{DA-Cal (ours)} &\textbf{0.0376} & \textbf{0.4835}& \textbf{0.3116}& \textbf{0.0325}& \textbf{0.3836}& \textbf{0.2324} \\
    \hline
\rowcolor{yellow!5}     Oracle &0.0510& 0.4019&0.2410& 0.0510& 0.4019&0.2410 \\
     
        \bottomrule
    \end{tabular}
      \vspace{-1em}
\end{table}

\begin{table}[t]

\small
    \centering
    \caption{Segmentation Performance  (mIoU) with DA-Cal. $*$ denotes the reproduced result.} \label{table3}
    \setlength{\tabcolsep}{3pt}
    \begin{tabular}{l|cc|cc}
     \toprule
      \textbf{Method} 
        &baseline &w/ DA-Cal  &baseline &w/ DA-Cal  
         \\
             \hline 
   & \multicolumn{2}{c|}{\textbf{GTA.$\rightarrow$CS.}} & \multicolumn{2}{c}{\textbf{SYN.$\rightarrow$CS.}}  \\ 
      \hline 
   DACS* \cite{tranheden2021dacs}& 51.9 & \cellcolor{blue!5}  \textbf{53.5}& 47.9&    \cellcolor{blue!5} \textbf{49.7} \\
    DAFormer \cite{hoyer2022daformer} & 68.3 & \cellcolor{blue!5} \textbf{69.4} & 60.9&   \cellcolor{blue!5} \textbf{62.2}\\
    MIC* \cite{hoyer2023mic}& 75.5& \cellcolor{blue!5}  \textbf{76.3}&67.1 &  \cellcolor{blue!5} \textbf{68.0}  \\
    \hline
   & \multicolumn{2}{c|}{\textbf{VNC. $\rightarrow$ Lucchi}} & \multicolumn{2}{c}{\textbf{Mito.-R $\rightarrow$ Mito.-H}}  \\ 
      \hline 
  DA-ISC \cite{huang2022domain} & 68.7& \cellcolor{blue!5} \textbf{70.5} & 74.8&   \cellcolor{blue!5} \textbf{76.1} \\
  CAFA \cite{yin2023class} &71.8 & \cellcolor{blue!5} \textbf{72.9} &76.3 & \cellcolor{blue!5}  \textbf{77.0}\\
        \bottomrule
    \end{tabular}
     \vspace{-1em}
\end{table}

\subsection{Results}

\noindent \textbf{Calibration Performance.}
DA-Cal significantly improves calibration across all benchmarks and UDA methods.
In autonomous driving scenarios (Tab. \ref{table1}), our method reduces ECE from 10.42\% to 4.59\% with MIC on GTA5$\rightarrow$Cityscapes—a 56\% relative improvement—alongside substantial reductions in NLL (29\%) and BS (30\%). Similar improvements are observed for SYNTHIA$\rightarrow$Cityscapes and Cityscapes$\rightarrow$ACDC, where DA-Cal consistently outperforms all calibration baselines. These gains can be attributed to our pixel-wise MTN, which effectively captures spatial variations in prediction uncertainty, and the bi-level optimization framework, which facilitates more effective transfer of calibration knowledge from source to target domains.

In biomedical imaging scenarios (Tab. \ref{table2}), where domain shifts are more pronounced, DA-Cal achieves even greater improvements. For VNC III$\rightarrow$Lucchi, our method reduces ECE from 21.67\% to 5.20\% with CAFA—a 76\% reduction—with similarly strong results for MitoEM-R$\rightarrow$MitoEM-H (77\% ECE reduction). These results confirm the robustness of DA-Cal in scenarios with severe domain shifts.

\noindent \textbf{Segmentation Performance.}
In addition to improving calibration, DA-Cal consistently enhances segmentation accuracy across all benchmarks (Tab. \ref{table3}).
For autonomous driving, our method delivers steady mIoU gains over all baselines: 1.6\%, 1.1\%, and 0.8\% improvements for DACS, DAFormer, and MIC on GTA5$\rightarrow$Cityscapes, with even larger gains on SYNTHIA$\rightarrow$Cityscapes (1.8\%, 1.3\%, and 0.9\%).
In biomedical imaging, DA-Cal achieves notable performance boosts: 1.8\% and 1.1\% mIoU improvements for VNC III$\rightarrow$Lucchi with DA-ISC and CAFA, and 1.3\% and 0.7\% for MitoEM-R$\rightarrow$MitoEM-H.
These consistent improvements demonstrate that integrating both hard and calibrated soft pseudo-labels into the unsupervised loss preserves valuable uncertainty information that would otherwise be discarded, effectively reducing confirmation bias in self-training.
By simultaneously improving both calibration and segmentation performance, DA-Cal highlights the intrinsic connection between reliable confidence estimation and accurate semantic prediction in domain adaptation.

\subsection{Ablation Study and Discussion}
Additional analyses can be found in Appendix  \ref{sec:D} to \ref{sec:H}.

\begin{table}[t]
\centering
\caption{Analysis of Different Implementation with
 DAFormer\cite{hoyer2022daformer} on GTAv $\rightarrow$Cityscapes.}
\tabcolsep=3.5pt
\label{table4}
\small
\begin{tabular}{cccc|cc} 
    \hline \rule{0pt}{1pt} % 增大高度
     ${\mathcal{L}}_{hard}$  & ${\mathcal{L}}_{soft}$ &
         DA-Cal-PH&  DA-Cal-BI& mIoU ($\uparrow$)&ECE  ($\downarrow$) \\ 
      \hline \hline  
     \checkmark & &  &  &68.3& 0.1255\\ 
     
    & \checkmark& &  &61.3&0.0857 \\ 
      \rowcolor{red!5}         &  &  \checkmark& &67.5& 0.0681\\
    
     \checkmark &  \checkmark& && 68.6& 0.1048 \\
    \rowcolor{blue!5}      \checkmark & & \checkmark &&\textbf{69.4}&\textbf{0.0578}\\ 
    \rowcolor{blue!5}     \checkmark & && \checkmark &69.1&0.0603\\ 
    \hline
    \end{tabular}
  \vspace{-1em}
\end{table}

\noindent \textbf{Analysis of Implementation.}
Tab.~\ref{table4} compares different implementation strategies on DAFormer~\cite{hoyer2022daformer}. 
From a theoretical perspective (Sec.~\ref{3.2}), perfectly calibrated soft pseudo-labels are equivalent to hard labels in expectation, as they reflect the true class likelihoods. Intuitively, soft labels act as a weighted average over all possible classes, and when properly calibrated, they provide the same expected supervision as stochastic sampling from hard labels.
Empirically, however, we observe that uncalibrated soft labels significantly degrade segmentation performance, even though they yield better calibration. For instance, replacing hard labels with soft ones (Row~2) causes a 7.0\% drop in mIoU, despite a 0.0398 improvement in ECE. This highlights that while soft labels are theoretically sound, their effectiveness in practice is heavily dependent on calibration quality. 
{\color{black}Moreover, this observation can be interpreted via the contrapositive of Sec.~\ref{3.2}: since perfect calibration is a sufficient condition under which soft/hard pseudo-label training becomes equivalent in expectation, a clear soft--hard performance gap implies that the sufficient condition is violated in practice, i.e., the soft pseudo-labels are not perfectly calibrated; thus miscalibration is a dominant contributor to the degradation rather than an inherently weaker expected supervision signal.}
In \colorbox{red!5}{Row 3}, we isolate the effect of using only calibrated soft labels with our DA-Cal-PH variant. Although calibration improves (ECE drops to 0.0681), segmentation performance (67.5\% mIoU) still lags behind the hard-label baseline (68.3\% mIoU). {\color{black}This indicates a small residual gap attributable to different optimization dynamics/error-propagation behaviors between soft and hard supervision, even after mitigating miscalibration, and suggests that soft-only supervision can be less stable in early training when model confidence is unreliable.}
 In contrast, hard labels—despite introducing confirmation bias—provide clearer and more consistent learning signals.
When combining both types of labels (Row~4), we observe modest improvements in both segmentation (+0.3\% mIoU) and calibration (-0.0207 ECE), highlighting the complementary effect of these supervision signals.
The introduction of our DA-Cal framework (\colorbox{blue!5}{Rows 5/6}) delivers substantial improvements. With DA-Cal-PH, we achieve a remarkable calibration improvement (0.0677 ECE reduction) alongside a significant segmentation performance gain (+1.1\% mIoU). Similarly, DA-Cal-BI yields comparable improvements (0.0652 ECE reduction and +0.8\% mIoU gain), with the added benefit of introducing no additional inference overhead. These results demonstrate that well-calibrated soft pseudo-labels not only improve confidence estimation but also enhance segmentation accuracy by providing more reliable supervision signals during domain adaptation.

\begin{figure}
    \centering
    \includegraphics[width=1\linewidth]{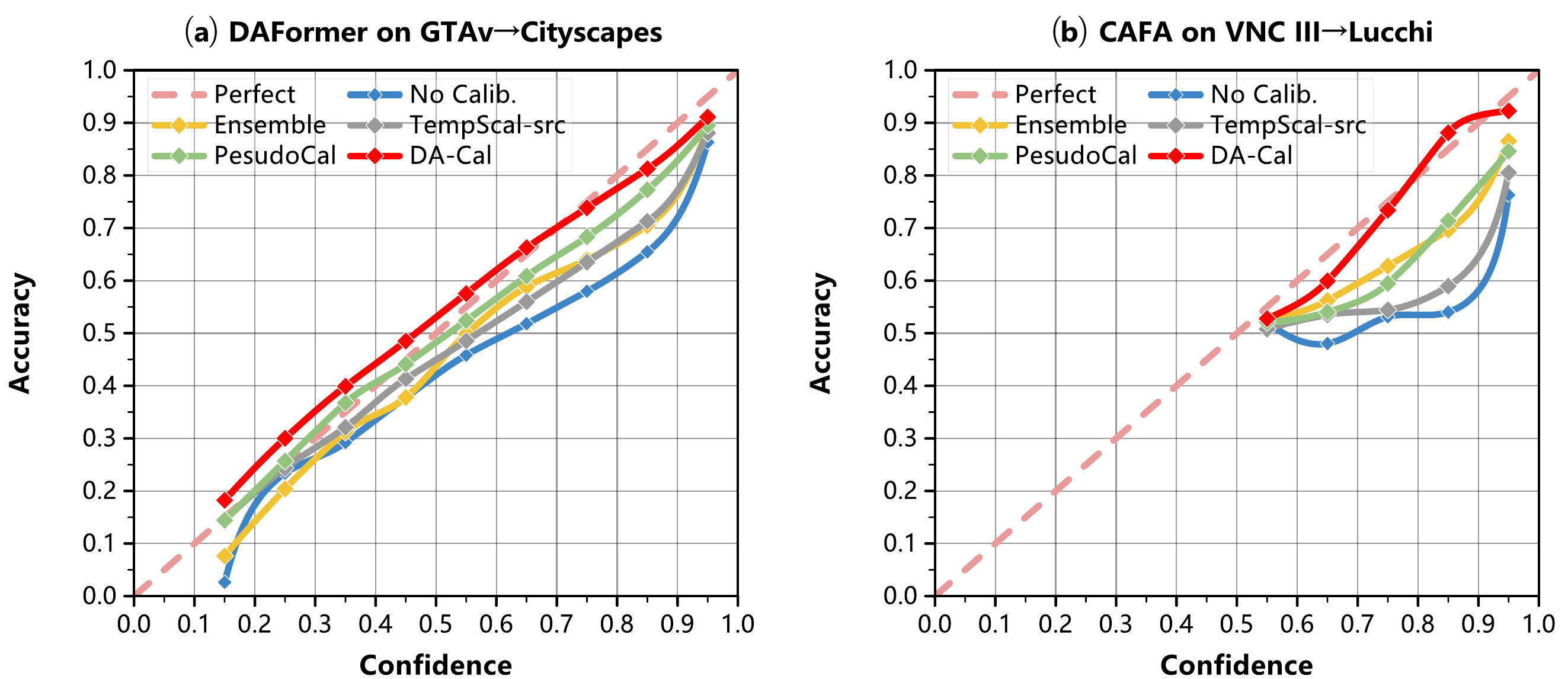}
    \caption{Comparison of DA-Cal and Baseline Methods on Reliability Diagrams.}
    \label{fig3}
    \vspace{-1em}
\end{figure}

\noindent \textbf{Qualitative Comparisons.}
Fig. \ref{fig3} clearly demonstrates our method's superior calibration performance compared to baseline approaches. The reliability diagrams show that our approach nearly perfectly aligns with the ideal diagonal line (representing perfect calibration), with minimal gaps between predicted confidence and actual accuracy across all confidence bins. While other methods exhibit significant calibration errors, our method achieves remarkable calibration precision, resulting in confidence scores that accurately reflect true prediction reliability in cross-domain settings. 

\begin{figure}
    \centering
    \includegraphics[width=1\linewidth]{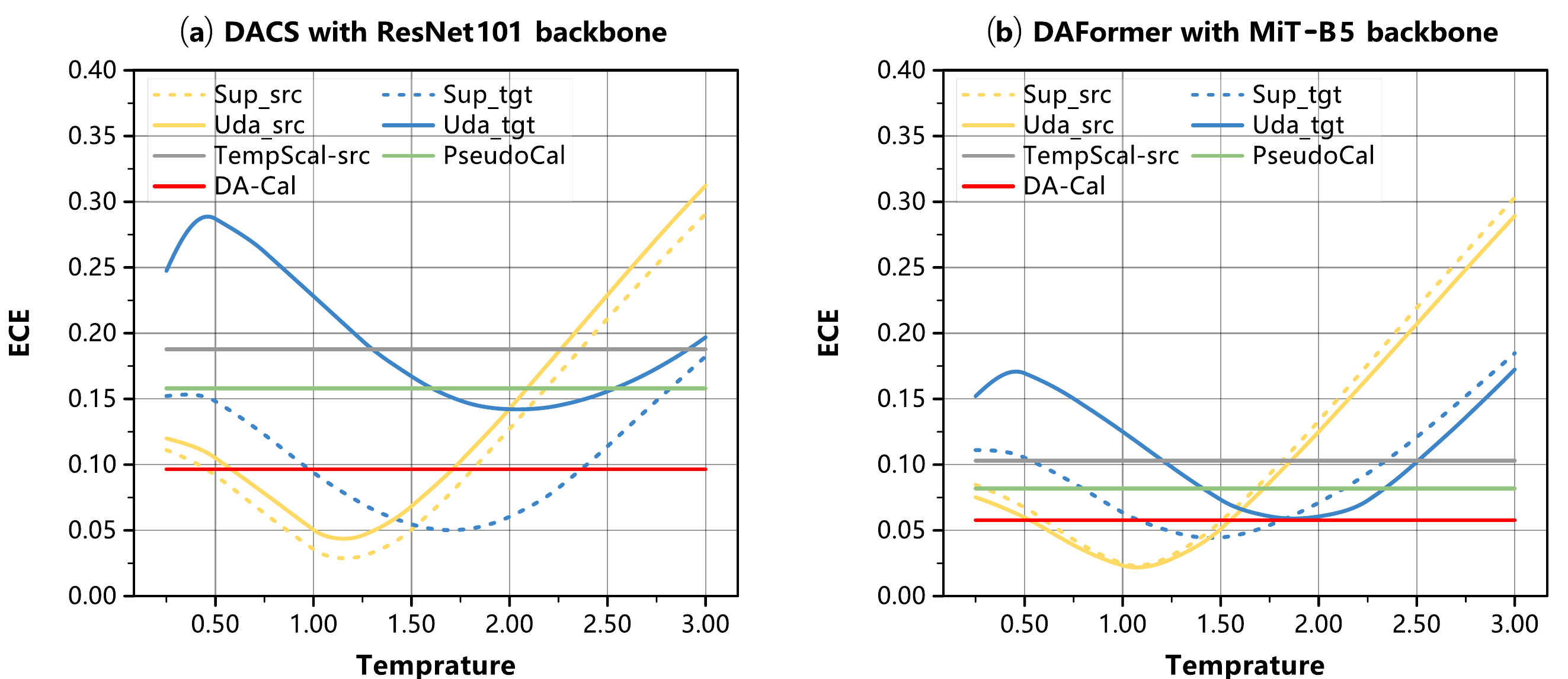}
    \caption{Ablation study on temperature scaling for GTAv. $\rightarrow$CS. benchmark across different methods.}
    \label{fig4}
     \vspace{-1em}
\end{figure}

\noindent \textbf{Discussion on Temperature Scaling.}
Fig. \ref{fig4} demonstrates the relationship between calibration performance and temperature values across different methods. We observe that source and target domains exhibit notably different optimal temperature parameters—a challenge further amplified in domain adaptation settings. This discrepancy means that temperatures optimized on source data (TempScal-src) yield suboptimal calibration when applied to target domains.
While PseudoCal \cite{hu2024pseudo} attempts to address this through pseudo-target synthesis, its sample-level approach is ill-suited for the pixel-level nature of segmentation tasks. Furthermore, conventional temperature scaling applies a global parameter without considering the crucial spatial context in segmentation.
Our DA-Cal method (red line) maintains superior calibration across all temperature values and network architectures by explicitly modeling target domain calibration with spatial awareness. It also eliminates the need for temperature parameter tuning, offering a robust solution when target domain labels are unavailable for calibration optimization.

\begin{figure}
    \centering
    \includegraphics[width=1\linewidth]{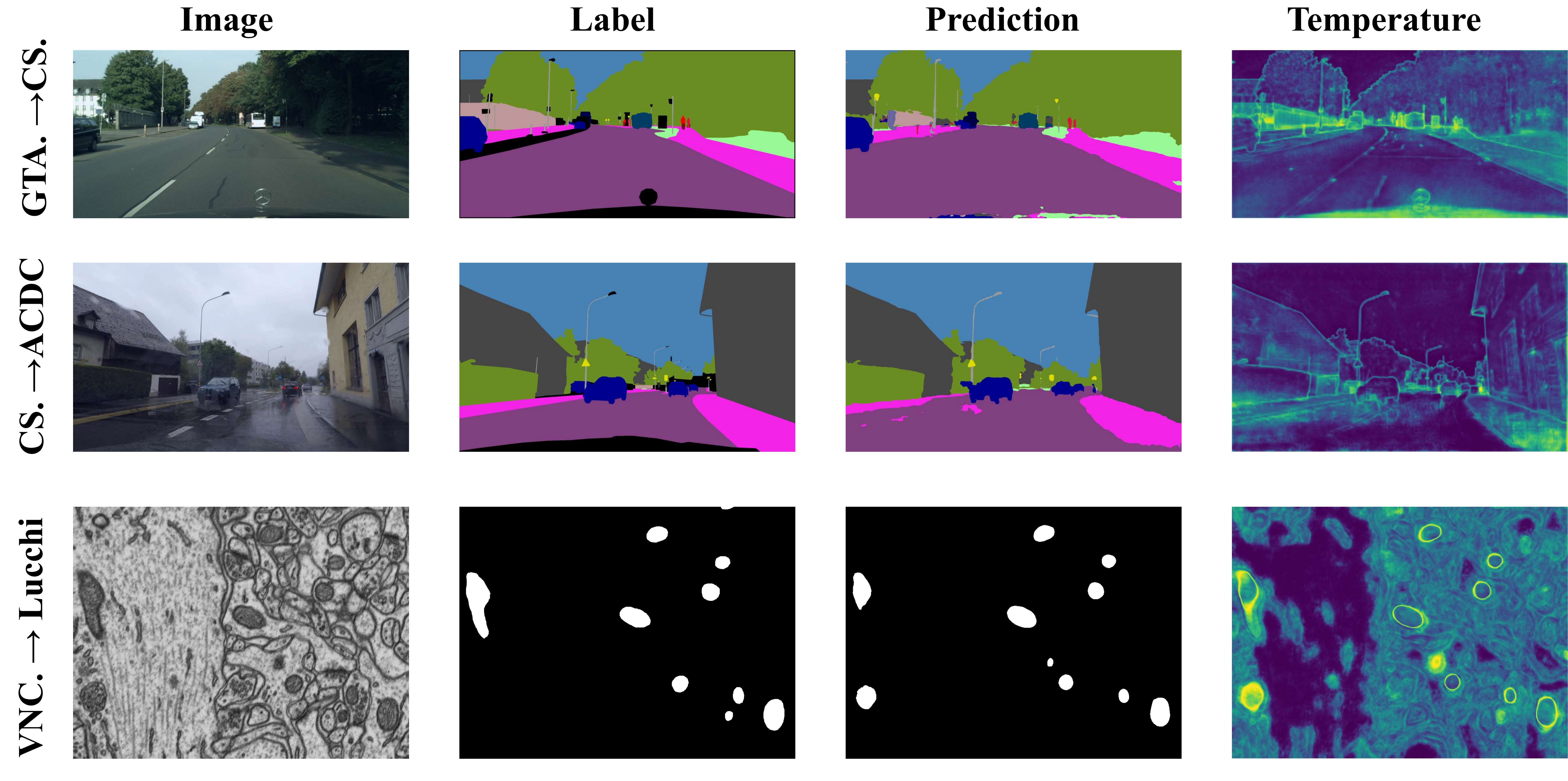}
    \caption{ Visualization  of temperature maps on different benchmarks. }
    \label{fig5}
      \vspace{-1em}
\end{figure}
\noindent \textbf{Visualization of Temperature Maps.}
Fig. \ref{fig5} illustrates the learned temperature maps across various domain adaptation benchmarks. These visualizations highlight how DA-Cal adaptively calibrates model confidence in a spatially-aware manner. The temperature maps serve a dual purpose: they sharpen logit distributions in regions with reliable predictions (indicated by lower $\mathcal{T}$ values and darker colors), while smoothing distributions in potentially erroneous areas (higher $\mathcal{T}$ values with lighter colors).
This spatial adaptability is especially beneficial for semantic segmentation, where prediction confidence can vary significantly across different image regions. Lower temperatures help preserve confident predictions in reliable areas, whereas higher temperatures suppress overconfident yet incorrect predictions in uncertain regions. This dynamic calibration approach greatly improves the model’s reliability across diverse domain adaptation scenarios.

\section{Conclusion}
In this work, we present DA-Cal, a novel cross-domain calibration framework designed for unsupervised domain adaptation (UDA) in semantic segmentation. Our method reformulates the target domain calibration challenge as an optimization problem over soft pseudo-labels, leveraging a bi-level optimization strategy inspired by meta-learning. At its core, DA-Cal introduces the Meta Temperature Network (MTN), which enables pixel-wise adaptive temperature scaling, effectively reducing calibration errors that hinder UDA performance. To further boost generalization and prevent overfitting, we incorporate complementary domain-mixing strategies. DA-Cal integrates seamlessly into existing self-training UDA frameworks, offering improved calibration quality and enhanced segmentation accuracy—all without introducing additional inference overhead. Extensive experiments on multiple benchmarks demonstrate the effectiveness of our approach, establishing a new standard for reliable and well-calibrated cross-domain semantic segmentation.

\section{APPENDIX}

This appendix provides additional details and analyses to complement our main paper. We include in-depth explanations of our evaluation metrics, implementation details, ablation studies, and additional visualization results. 
% Specifically, the appendix is organized as follows:
% \begin{itemize}
%     \item \textbf{Section~\ref{sec:A}} introduces the calibration metrics used in our evaluation, including Expected Calibration Error (ECE), Negative Log-Likelihood (NLL), and Brier Score (BS). We also describe how these metrics are adapted for semantic segmentation.
%     \item \textbf{Section~\ref{sec:B}}  provides details on the baseline calibration methods used for comparison.
%     \item \textbf{Section~\ref{sec:C}}  describes additional implementation details of DA-Cal, including the design of the Meta Temperature Network (MTN) and optimization strategies.
%     \item \textbf{Section~\ref{sec:D}}  presents an analysis of training strategies, demonstrating the effectiveness of EMA-based MTN updates and warm-up calibration.
%     \item \textbf{Section~\ref{sec:E}}  explores the impact of DA-Cal’s hyperparameters on performance.
%     \item \textbf{Section~\ref{sec:F}}  evaluates the computational efficiency of DA-Cal in terms of GPU memory and training time.
%     \item \textbf{Section~\ref{sec:G}}  studies the effect of different domain-mixing strategies on segmentation and calibration performance.
%     \item \textbf{Section~\ref{sec:H}}  showcases additional visualization results to further illustrate the effectiveness of DA-Cal.

% \end{itemize}

{\color{black}
\subsection{Bi-level Optimization Formulation of DA-Cal}
\label{sec:bilevel_formulation}

We provide a principled view of DA-Cal as a bi-level optimization problem and explain why the meta-update in Eq.~\eqref{eq:13} is well-defined. Let \(f_\theta\) denote the student network and \(h_\psi\) denote MTN. In each iteration \(k\), we create a temporary copy \(\theta'^{[k]}\leftarrow \theta^{[k]}\) and perform an \emph{inner} update of \(\theta'\) using calibrated soft supervision (Step 1). Denote the inner objective by \(\mathcal{L}_{\text{cal}}(S,T;\theta',\psi)\) (as described around Eq.~\eqref{eq:11}); the one-step inner update is
\begin{equation}
\theta'^{[k+1]} = \theta'^{[k]} - \alpha \nabla_{\theta'^{[k]}} \mathcal{L}_{\text{cal}}(S,T;\theta'^{[k]},\psi^{[k]}),
\label{eq:app_inner_theta_update}
\end{equation}
which matches Eq.~\eqref{eq:11}. This update makes \(\theta'^{[k+1]}\) an explicit function of \(\psi^{[k]}\) through \(\mathcal{L}_{\text{cal}}\), because MTN affects the calibrated probabilities used to compute \(\mathcal{L}_{\text{cal}}\).

To determine whether a calibration update is truly beneficial for UDA, we define an \emph{outer} objective on complementary mixed data (Step 2). Let \(\mathcal{L}_{\text{mix}}(\theta')\) denote the cross-entropy loss evaluated on the ClassMix-constructed mixed samples (with mixed supervision as described under Eq.~\eqref{eq:13}). The bi-level objective is then
\begin{equation}
\min_{\psi}\ \mathcal{L}_{\text{mix}}(\theta'^{[k+1]})
\quad \text{s.t.}\quad
\theta'^{[k+1]} \ \text{given by Eq.~\eqref{eq:app_inner_theta_update}}.
\label{eq:app_bilevel_objective}
\end{equation}

Since \(\mathcal{L}_{\text{cal}}\) is differentiable in \((\theta',\psi)\) and \(\mathcal{L}_{\text{mix}}\) is differentiable in \(\theta'\), the outer objective \(\mathcal{L}_{\text{mix}}(\theta'^{[k+1]}(\psi))\) is a differentiable composite function of \(\psi\). By the chain rule, the meta-gradient used to update MTN is
\begin{equation}
\begin{aligned}
\nabla_{\psi^{[k]}} &\mathcal{L}_{\text{mix}}(\theta'^{[k+1]})
=
\frac{\partial \mathcal{L}_{\text{mix}}}{\partial \theta'^{[k+1]}}
\cdot
\frac{\partial \theta'^{[k+1]}}{\partial \psi^{[k]}} \\
&=
-\alpha
\frac{\partial \mathcal{L}_{\text{mix}}}{\partial \theta'^{[k+1]}}
\cdot
\frac{\partial}{\partial \psi^{[k]}}
\Bigl(\nabla_{\theta'^{[k]}} \mathcal{L}_{\text{cal}}(S,T;\theta'^{[k]},\psi^{[k]})\Bigr),
\end{aligned}
\label{eq:app_meta_gradient}
\end{equation}
which corresponds exactly to the implementation in Eq.~\eqref{eq:13} via automatic differentiation. Intuitively, MTN is encouraged to produce calibrated probabilities that lead (after the inner update) to a student model \(\theta'\) with lower loss on complementary mixed data, i.e., calibration is optimized only insofar as it improves downstream self-training.

In practice, Eq.~\eqref{eq:app_meta_gradient} can be computed with full second-order differentiation (as written), and a first-order approximation can also be used by ignoring higher-order terms to reduce computation, following common meta-learning practice.}

\subsection{Introduction of Calibration Metrics}
\label{sec:A}

In this section, we introduce the calibration metrics used in our evaluation, including Expected Calibration Error (ECE), Negative Log-Likelihood (NLL), and Brier Score (BS). 

\textbf{Expected Calibration Error (ECE) \cite{guo2017calibration}} measures the difference between the predicted confidence and the actual accuracy. It is computed by partitioning predictions into multiple bins and calculating the weighted average of absolute differences between confidence and accuracy in each bin. The formula is defined as:
\begin{equation}
    \text{ECE} = \sum_{m=1}^{M} \frac{|B_m|}{N} \left| \text{acc}(B_m) - \text{conf}(B_m) \right|,
\end{equation}
where $M$ is the number of bins, $B_m$ denotes the set of predictions that fall into the $m$-th bin, $|B_m|$ is the number of samples in that bin, $N$ is the total number of samples, $\text{acc}(B_m)$ is the average accuracy of samples in bin $m$, and $\text{conf}(B_m)$ is the average confidence.

\textbf{Negative Log-Likelihood (NLL) \cite{goodfellow2016deep}} is a well-known metric for assessing probabilistic predictions, which is also known as the cross-entropy loss. It evaluates how well a model assigns probabilities to the correct labels, with lower values indicating better calibration. The NLL is defined as:
\begin{equation}
    \text{NLL} = - \frac{1}{N} \sum_{i=1}^{N} \sum_{c=1}^{C} y_{i,c} \log p_{i,c},
\end{equation}
where $N$ is the total number of pixels, $C$ is the number of classes, $y_{i,c}$ is the one-hot encoded ground-truth label indicating whether pixel $i$ belongs to class $c$, and $p_{i,c}$ is the predicted probability for class $c$.

\textbf{Brier Score (BS) \cite{brier1950verification}} quantifies the mean squared error between predicted probabilities and the actual one-hot encoded labels. It captures both calibration and sharpness aspects of the predictions. The BS is defined as:
\begin{equation}
    \text{BS} = \frac{1}{N} \sum_{i=1}^{N} \sum_{c=1}^{C} (p_{i,c} - y_{i,c})^2.
\end{equation}

Unlike classification tasks, where these metrics are computed on a sample basis, semantic segmentation involves per-pixel evaluation, leading to severe class imbalance issues. To address this, we compute ECE, NLL, and BS at the class level and then take an average across all classes:
\begin{equation}
    \mathcal{M}_{\text{class}} = \frac{1}{C} \sum_{c=1}^{C} \mathcal{M}(c),
\end{equation}
where $\mathcal{M}$ represents any of the above metrics (ECE, NLL, or BS) computed for class $c$. 

To improve computational efficiency, we sample 10,000 pixels per image to estimate these metrics following \cite{de2023reliability}.

\subsection{Details of Calibration Baselines}
\label{sec:B}

We compare DA-Cal with several existing calibration methods:

\textbf{Ensemble \cite{mehrtash2020confidence, lakshminarayanan2017simple}:} We train each segmentation model three times with different random seeds. During inference, we average the predicted logits from these models before applying softmax to obtain final calibrated probabilities.

{\color{black}\textbf{MC Dropout~\cite{gal2016dropout}:} We include MC Dropout as a standard lightweight uncertainty-estimation baseline. We keep dropout layers enabled at inference and perform \(T=5\) stochastic forward passes for each input. We then average the resulting logits and apply softmax to obtain the final predictive probabilities. This baseline increases test-time computation proportionally to \(T\), and therefore serves as a cost-aware reference compared with larger deep ensembles.}

\textbf{TempScal-src \cite{guo2017calibration}:} This method applies temperature scaling using a hold-out validation set from the source domain to find the optimal temperature $\mathcal{T}$ to minimize NLL. The calibrated probabilities are computed as:
\begin{equation}
    p_{\text{cal}} = \text{softmax}(z/\mathcal{T}),
\end{equation}
where $z$ represents the pre-softmax logits.

{\color{black}\textbf{TempScal-src-LTS \cite{ding2021local}:} We further include a segmentation-specific post-hoc calibration baseline by applying Local Temperature Scaling (LTS), which generalizes temperature scaling to spatially varying temperatures for dense prediction. Specifically, LTS learns a local temperature field \(\mathcal{T}(x)\)  from the source-domain validation set by minimizing NLL, and calibrates each pixel as \(p_{\text{cal}}(x)=\text{softmax}(z(x)/\mathcal{T}(x))\).}

\textbf{PseudoCal \cite{hu2024pseudo}:} This method synthesizes pseudo-target samples using a mixup \cite{zhang2017mixup} strategy. The synthesized samples are used to learn temperature scaling parameters in the target domain. Unlike classification-based approaches that model samples independently, we apply PseudoCal at the pixel level.

{\color{black}\textbf{PseudoCal-LTS:} In addition, we combine PseudoCal with LTS by learning local (pixel-wise) temperatures from the pseudo-target (mixup-synthesized) data, and then applying the resulting \(\mathcal{T}(x)\) for pixel-wise calibration, analogous to TempScal-src-LTS but using pseudo-target supervision for calibration.}

\textbf{NoAdapt/Oracle:} The NoAdapt baseline trains models only on the source domain and directly evaluates them on the target domain. The Oracle model, on the other hand, is trained with labels from the target domain, providing an well-calibrated baseline for calibration performance.

\subsection{More Implementation Details of DA-Cal}
\label{sec:C}

\noindent\textbf{EMA Update and Warm-up Strategy:}  
The Meta Temperature Network (MTN) parameters are updated using an Exponential Moving Average (EMA) strategy. EMA ensures stable updates by gradually incorporating new knowledge while retaining past information. Since MTN progressively enhances calibration performance, we introduce a warm-up strategy to gradually increase the weight of soft pseudo-label supervision:
\begin{equation}
    \lambda_{\text{soft}} = \min(1, \frac{t}{T_{\text{warm-up}}}),
\end{equation}
where $t$ is the current training iteration, and $T_{\text{warm-up}}$ is the total number of warm-up iterations, which we set to half of the total training iterations. A detailed analysis of the effectiveness of the EMA update and warm-up strategy is provided in Section~\ref{sec:D}.

\noindent\textbf{MTN Architecture:}  
The MTN is designed as a lightweight calibration module to minimize computational overhead. It consists of a stack of convolutional layers followed by BatchNorm and ReLU activation. To balance model complexity and performance, we adopt a three-layer architecture in our experiments. A comprehensive analysis of the trade-off between MTN capacity and performance is provided in Section~\ref{sec:E}.

\noindent\textbf{Efficient Inner Optimization:}  
To improve computational efficiency, we optimize only the segmentation head while keeping the backbone frozen during inner-loop updates. This significantly reduces the memory footprint and accelerates training. The high-order meta-loss is computed solely for the segmentation head parameters, ensuring efficient parameter updates. A detailed evaluation of the computational overhead introduced by DA-Cal, including GPU memory usage and training time, is presented in Section~\ref{sec:F}

\subsection{Training Strategies Analysis of DA-Cal}
\label{sec:D}

We evaluate the effectiveness of EMA-based MTN updates and warm-up calibration for soft pseudo-labels on the DAFormer \cite{hoyer2022daformer} using the GTAv$\rightarrow$Cityscapes benchmark. Table~\ref{tab:ema_warmup} reports the impact of these strategies on segmentation and calibration performance.

\begin{table}[t]
    \centering \tabcolsep=3.0pt
        \caption{Impact of EMA updates and warm-up calibration on DA-Cal performance.}
    \begin{tabular}{ccccc} 
        \toprule
   Base & EMA(MTN) & Warm-up ($\lambda_{\text{soft}}$) & mIoU (\%) ↑ & ECE ↓ \\
        \midrule
     
 \rowcolor{black!5}  \checkmark   & -&-    & 68.3& 0.1255 \\ \hline
   - & -&- & 68.7&0.0687\\
-   &     \checkmark&- &  68.9& 0.0591 \\
    -   &  -& \checkmark & 69.2 & 0.0633 \\
  \rowcolor{blue!5}    - &   \checkmark &\checkmark  & \textbf{69.4} & \textbf{0.0578} \\
        \bottomrule
    \end{tabular}

    \label{tab:ema_warmup}
    \vspace{-1em}
\end{table}

\noindent\textbf{Effect of Warm-up Strategy on Segmentation Performance.}
From Table~\ref{tab:ema_warmup}, we observe that applying the warm-up strategy ($\lambda_{\text{soft}}$) significantly improves the segmentation performance (mIoU). Specifically, compared to the baseline without warm-up (row 2 vs. row 4), mIoU increases from 68.7\% to 69.2\%, demonstrating a clear benefit. This improvement can be attributed to the gradual introduction of soft pseudo-label supervision, which prevents early-stage noisy pseudo-labels from negatively affecting model training. By progressively increasing the weight of soft pseudo-labels, the network learns more reliable supervision signals, thereby enhancing segmentation accuracy.

\noindent\textbf{Effect of EMA-based MTN Updates on Calibration Performance.}
The results also indicate that the EMA-based MTN updates primarily contribute to improving calibration performance. When enabling EMA (row 2 vs. row 3), the Expected Calibration Error (ECE) decreases from 0.0687 to 0.0591, showing that the EMA strategy effectively stabilizes temperature updates, leading to better calibration. This is because EMA ensures smoother updates to the MTN parameters, preventing abrupt fluctuations and improving the consistency of pixel-wise temperature scaling.

Furthermore, combining EMA with the warm-up strategy (row 5) achieves the best overall performance, with mIoU = 69.4\% and ECE = 0.0578. This confirms that while warm-up enhances segmentation accuracy, EMA plays a crucial role in refining the calibration quality.

\subsection{Hyper-parameters Analysis}
\label{sec:E}

The key hyperparameters of DA-Cal include the number of layers in MTN, the learning rate for inner optimization, and the learning rate for MTN. We use SGD with a learning rate of 0.01 for both inner optimization and MTN training. Table~\ref{tab:hyperparam} explores the impact of these hyperparameters wtih DAFormer \cite{hoyer2022daformer}  using the GTAv$\rightarrow$Cityscapes benchmark.

\begin{table}[t]
    \centering \tabcolsep=6pt    \caption{Hyperparameter sensitivity analysis for DA-Cal.}
\begin{tabular}{c|cccc} 
\hline
MTN layers& 1 & 2& 3 &4\\
\hline
mIoU (\%) ↑& 68.9&69.4 & \cellcolor{blue!5} \textbf{69.4} & 69.3 \\
 ECE ↓ &0.0672 &0.0610&\cellcolor{blue!5} 0.0578 & \textbf{0.0569} \\
\hline
\hline
Inner LR& 0.1 & 0.05& 0.01 &0.001\\
\hline
mIoU (\%) ↑& 68.7& 69.4 & \cellcolor{blue!5} \textbf{69.4}& 69.1 \\
 ECE ↓ & 0.0710 & 0.0593& \cellcolor{blue!5} \textbf{0.0578} & 0.0649 \\
\hline

\end{tabular}

    \label{tab:hyperparam}
     \vspace{-1em}
\end{table}

\noindent\textbf{Impact of MTN Depth on Performance.}
From Table~\ref{tab:hyperparam}, we observe that increasing the number of MTN layers from 1 to 3 improves both segmentation (mIoU) and calibration (ECE). Further increasing to 4 layers does not yield additional benefits. This suggests diminishing returns for deeper MTN architectures.
Based on these observations, we adopt a 3-layer MTN in our final model to balance calibration effectiveness and computational efficiency.

\noindent\textbf{Impact of Inner Learning Rate on Performance.}
For the inner optimization learning rate, we find that a learning rate of 0.01 yields the best results, while a higher learning rate (0.1) leads to degraded calibration performance (ECE = 0.0710), likely due to instability in meta-updates, and a lower learning rate (0.001) results in worse calibration (ECE = 0.0649) and slightly lower segmentation performance (69.1\% mIoU), suggesting slower convergence.  

\subsection{DA-Cal Efficiency Analysis}
\label{sec:F}

Our method is computationally efficient. Table~\ref{tab:efficiency} compares GPU memory usage and training time for DAFormer \cite{hoyer2022daformer} and DACS \cite{tranheden2021dacs} with and without DA-Cal.

\begin{table}[t]
    \centering  \caption{Comparison of GPU memory and training time.}
    \begin{tabular}{lcc}
        \toprule
        Method & GPU Memory (MB) & Training Time (h) \\
        \midrule
        DAFormer & 9,807 & 16.2 \\
    \rowcolor{blue!5}    w/ DA-Cal & 13,409 & 22.5 \\
        DACS & 11,078 & 25.4 \\
    \rowcolor{blue!5}     w/ DA-Cal & 14,485 & 31.9 \\
        \bottomrule
    \end{tabular}
   
    \label{tab:efficiency}
     \vspace{-1em}
\end{table}

\noindent\textbf{Computational Overhead Analysis.}  
From Table~\ref{tab:efficiency}, we observe that integrating DA-Cal introduces additional computational cost but remains practical. Specifically, for DAFormer, DA-Cal increases GPU memory usage from 9,807 MB to 13,409 MB (+36.7\%) and training time from 16.2 hours to 22.5 hours (+38.9\%). Similarly, for DACS, DA-Cal increases GPU memory from 11,078 MB to 14,485 MB (+30.7\%) and training time from 25.4 hours to 31.9 hours (+25.6\%).

\noindent\textbf{Efficiency-Performance Trade-off.}  
Although DA-Cal introduces moderate overhead, the significant improvements in both segmentation accuracy  and calibration quality (ECE) justify the additional cost. The increase in memory usage is mainly due to the meta-learning framework, which requires storing temporary network states for gradient computation. However, the efficient inner-loop optimization (Section~\ref{sec:C}) mitigates excessive computational burden by freezing the backbone, ensuring that the extra cost remains manageable.

\begin{table}[t]
\centering   \tabcolsep=1.0pt
\caption{Comparison of mixing strategies  with DAFormer on GTA $\rightarrow$ CS. and CAFA on VNC. $\rightarrow$ Lucchi.}
\label{tab:mixing}
 \begin{tabular}{c|c|cc|cc}
        \toprule
  \multicolumn{2}{c|}{\multirow{2}{*}{Mixing Strategy}}      &  \multicolumn{2}{c|}{\textbf{GTA $\rightarrow$ CS.}} & \multicolumn{2}{c|}{\textbf{VNC. $\rightarrow$ Lucchi}} \\  \cline{3-6}
  \multicolumn{2}{c|}{}  & mIoU $\uparrow$ & ECE $\downarrow$  & mIoU $\uparrow$ & ECE $\downarrow$\\
        \midrule

        Baseline&-&68.3&0.2289&71.8&0.2167\\ \hline
\multirow{3}{*}{\textbf{ClassMix}}  &    Same  & 69.0 & 0.0638  &72.0 &0.0785\\
       & Random& 69.3 & 0.0612 &72.0 &0.0774\\
   \rowcolor{blue!5}    & \textbf{Complementary} & \textbf{69.4} & \textbf{0.0578}  &72.1& 0.0705\\
   \hline
   \multirow{3}{*}{CutMix}  &    Same  &68.9 & 0.0641&72.6& 0.0596 \\
       & Random& 68.9 &0.0607 &72.9&0.0563  \\
   \rowcolor{blue!5}    & Complementary & 69.2 & 0.0591 & \textbf{72.9} & \textbf{0.0520} \\
        \bottomrule
    \end{tabular}
     \vspace{-1em}
\end{table}

\subsection{Ablation on Mixing Strategies}
\label{sec:G}

ClassMix \cite{olsson2021classmix} is used in our autonomous driving benchmarks, ensuring inner and outer optimization use complementary masks to avoid self-reinforcement and overfitting. Specifically, ClassMix randomly selects half of the source domain classes for mixing in each iteration. To enforce complementary learning, we assign the remaining half of the classes to the inner optimization, ensuring that the two optimization stages utilize disjoint class masks. This prevents the model from overfitting to a fixed subset of samples and improves generalization.

% \begin{table}[h]
%     \centering      \caption{Impact of mixing strategies on DA-Cal performance with DAFormer on GTA $\rightarrow$ CS.}
%     \begin{tabular}{lcc}
%         \toprule
%         Mixing Strategy & mIoU (\%) ↑ & ECE ↓ \\
%         \midrule
%         Same  & 69.0 & 0.0638 \\
%         Random& 69.3 & 0.0612\\
%    \rowcolor{blue!5}     Complementary  (Ours) & \textbf{69.4} & \textbf{0.0578} \\
%         \bottomrule
%     \end{tabular}

%     \label{tab:mixing}
% \end{table}

\noindent\textbf{Analysis of Mixing Strategies.}  
Table~\ref{tab:mixing} compares different mixing strategies. Using the same mask for both inner and outer optimization results in slightly lower performance (69.0\% mIoU, 0.0638 ECE), likely due to self-reinforcement effects where the model overfits to specific regions. Randomly selecting masks per iteration improves results (69.3\% mIoU, 0.0612 ECE), but lacks structured control over the mixed regions. Our complementary mask strategy further enhances performance (69.4\% mIoU, 0.0578 ECE) by ensuring that inner and outer optimization operate on disjoint class sets, leading to better generalization and calibration.

\begin{figure}[t]
    \centering
    \includegraphics[width=0.9\linewidth]{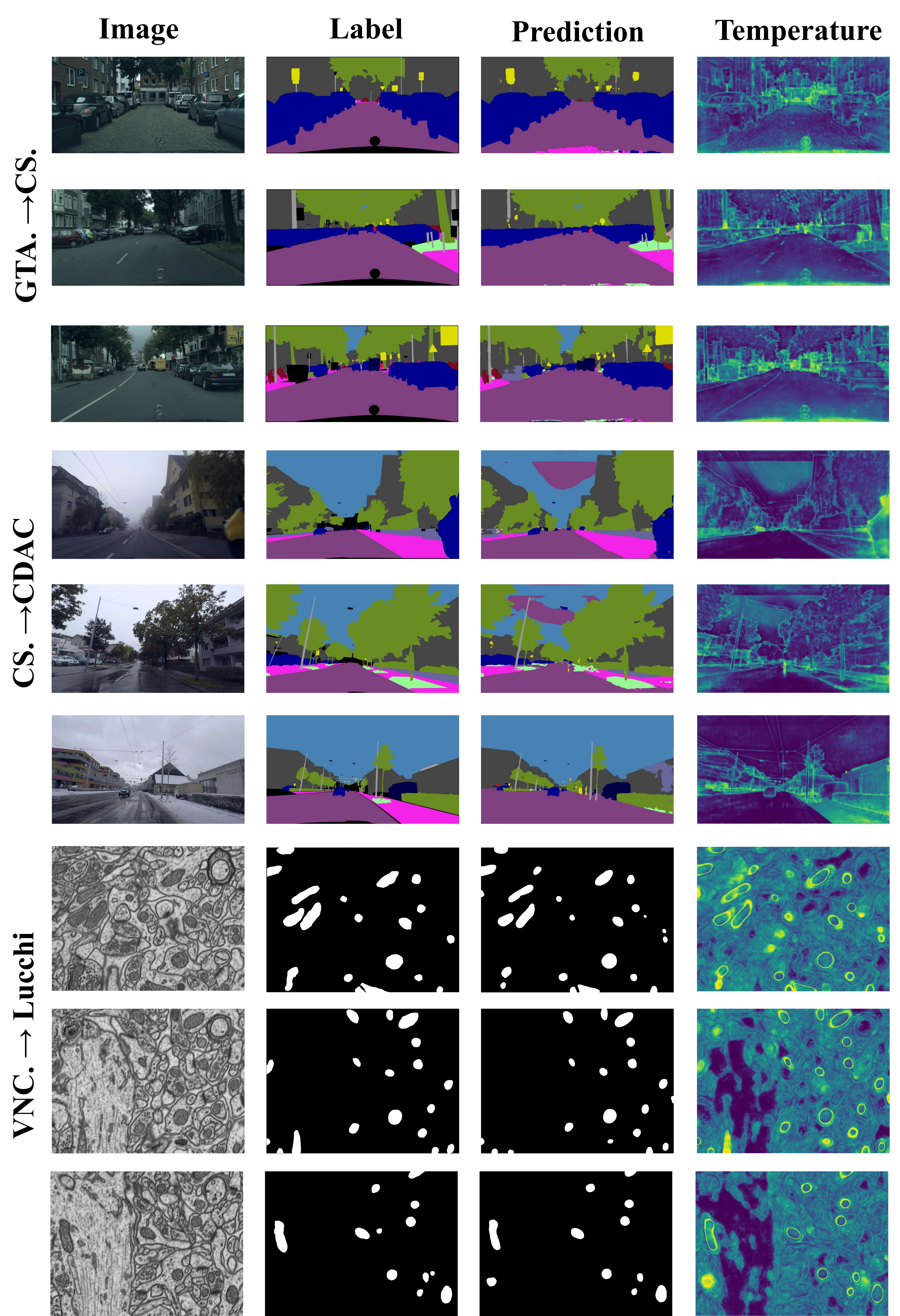}
    \caption{Visualization of learned temperature maps on different adaptation benchmarks: GTA. $\rightarrow$ CS. (synthetic-to-real), CS. $\rightarrow$ ACDC (clear-to-adverse weather), and VNC. $\rightarrow$ Lucchi (biomedical segmentation).}
    \label{fig:temp_maps}
     \vspace{-1em}
\end{figure}

\noindent\textbf{Region-Level CutMix for Biomedical Imaging.}  
For biomedical imaging, we adopt region-level CutMix \cite{yun2019cutmix}, which is more suitable for foreground segmentation tasks such as binary tissue segmentation. In binary settings like VNC $\rightarrow$ Lucchi, ClassMix becomes less effective due to the lack of class diversity. Specifically, ClassMix may remove all foreground regions during mixing, resulting in poor supervision signals.
Unlike ClassMix, which relies on class-wise separation and assumes multiple semantic categories, CutMix operates at the region level, directly mixing local patches from different images. This region-aware augmentation is particularly beneficial for binary tasks where the goal is to distinguish foreground structures from background.
To validate this, we compare the two strategies on this benchmark. As shown in Table~\ref{tab:mixing}, CutMix outperforms ClassMix, achieving a higher IoU (72.9\% vs. 72.1\%) and a lower expected calibration error (ECE: 0.0520 vs. 0.0704). These results highlight the importance of selecting augmentation strategies that align with the dataset characteristics—region-aware mixing proves essential for effective biomedical domain adaptation.

% \begin{table}[h]
%     \centering \caption{Comparison of ClassMix and CutMix on CAFA with VNC. $\rightarrow$ Lucchi benchmark.}
%     \begin{tabular}{lcc}
%         \toprule
%         Mixing Strategy & IoU (\%) ↑ & ECE ↓ \\
%         \midrule
%         ClassMix  & 72.1 & 0.0704 \\
%    \rowcolor{blue!5}     CutMix (Ours) & \textbf{72.9} & \textbf{0.0520}\\
%         \bottomrule
%     \end{tabular}
    
%     \label{tab:biomedical_mixing}
% \end{table}

\subsection{More Visualization Results}
\label{sec:H}
To further validate DA-Cal’s effectiveness, we provide additional temperature map visualizations across various domain adaptation tasks. Figure~\ref{fig:temp_maps} showcases results for synthetic-to-real adaptation (GTA. $\rightarrow$ CS.), clear-to-adverse weather adaptation (CS. $\rightarrow$ ACDC), and biomedical segmentation (VNC. $\rightarrow$ Lucchi). These results highlight how DA-Cal dynamically adjusts temperature values to enhance model calibration, particularly in regions prone to miscalibration.

\noindent\textbf{Syntheticto-Real Adaptation (GTA. $\rightarrow$ CS.).}  
For synthetic-to-real adaptation, we observe that regions affected by domain shifts, such as road textures, building facades, and vegetation, exhibit higher temperatures. This suggests that DA-Cal effectively identifies areas where synthetic textures differ from real-world counterparts, thereby reducing overconfidence in these regions. Moreover, object boundaries (e.g., vehicles, poles, pedestrians) tend to have elevated temperatures, preventing the model from being overly confident in ambiguous regions.

\noindent\textbf{Clear-to-Adverse Weather Adaptation (CS. $\rightarrow$ ACDC).}  
When adapting from clear-weather Cityscapes to the adverse-weather ACDC dataset, the temperature maps reveal higher uncertainty in regions affected by fog, snow, and rain. Road areas remain well-calibrated (low temperatures), while occluded objects and distant structures show elevated temperatures. This demonstrates that DA-Cal effectively captures visibility degradation, ensuring better calibration in uncertain regions.

\noindent\textbf{Biomedical Segmentation (VNC III$\rightarrow$ Lucchi).}  
In biomedical domain adaptation, the temperature maps exhibit greater adjustments around cellular boundaries, reflecting the inherent segmentation ambiguity in these regions. Unlike urban scenes—where uncertainty primarily arises from domain shifts—uncertainty here stems from the challenge of distinguishing fine-grained structures from background tissue. This suggests that DA-Cal enhances model reliability in complex foreground-background segmentation tasks, successfully adapting to the unique characteristics of biomedical imagery.

{\color{black}
\subsection{Domain-Specific Cue Analysis}
\label{sec:appendix_domain_disc}

This appendix empirically examines whether the proposed meta-transfer network (MTN) might exploit domain-specific cues due to taking \emph{image+logit} inputs, and whether our complementary mixing strategy sufficiently mitigates this risk.

\subsubsection{Setup: Domain discriminator with GRL}
\label{sec:appendix_domain_disc_setup}
We attach an auxiliary domain discriminator \(d_{\omega}\) \cite{ganin2015unsupervised} to the MTN features and train it to predict the domain label (source vs.\ target). Concretely, let \(\mathbf{u}\) denote the intermediate feature produced by MTN when processing a mixed sample (the same MTN parameters are applied to both source- and target-containing regions due to domain sharing). The discriminator outputs a domain probability \(\hat{r}=d_{\omega}(\mathbf{u})\in(0,1)\), where domain label \(r=1\) indicates source and \(r=0\) indicates target. The discriminator is optimized with the binary cross-entropy:
\begin{equation}
\mathcal{L}_{\text{dom}}(\omega) = - \mathbb{E}\big[r\log \hat{r} + (1-r)\log (1-\hat{r})\big].
\label{eq:dom_loss}
\end{equation}

To test whether MTN features contain domain-informative cues, we additionally employ a gradient reversal layer (GRL) between MTN and \(d_{\omega}\). With GRL enabled, the discriminator is still trained to minimize \(\mathcal{L}_{\text{dom}}\) w.r.t.\ \(\omega\), while MTN is trained to \emph{maximize} \(\mathcal{L}_{\text{dom}}\) w.r.t.\ \(\psi\) (via the reversed gradients). If MTN features are strongly domain-specific, enabling GRL should noticeably affect the discriminator's learning dynamics (e.g., increase its loss or prevent convergence), and may also influence the calibration behavior.

\begin{figure}[ht]
    \centering
    \includegraphics[width=1\linewidth]{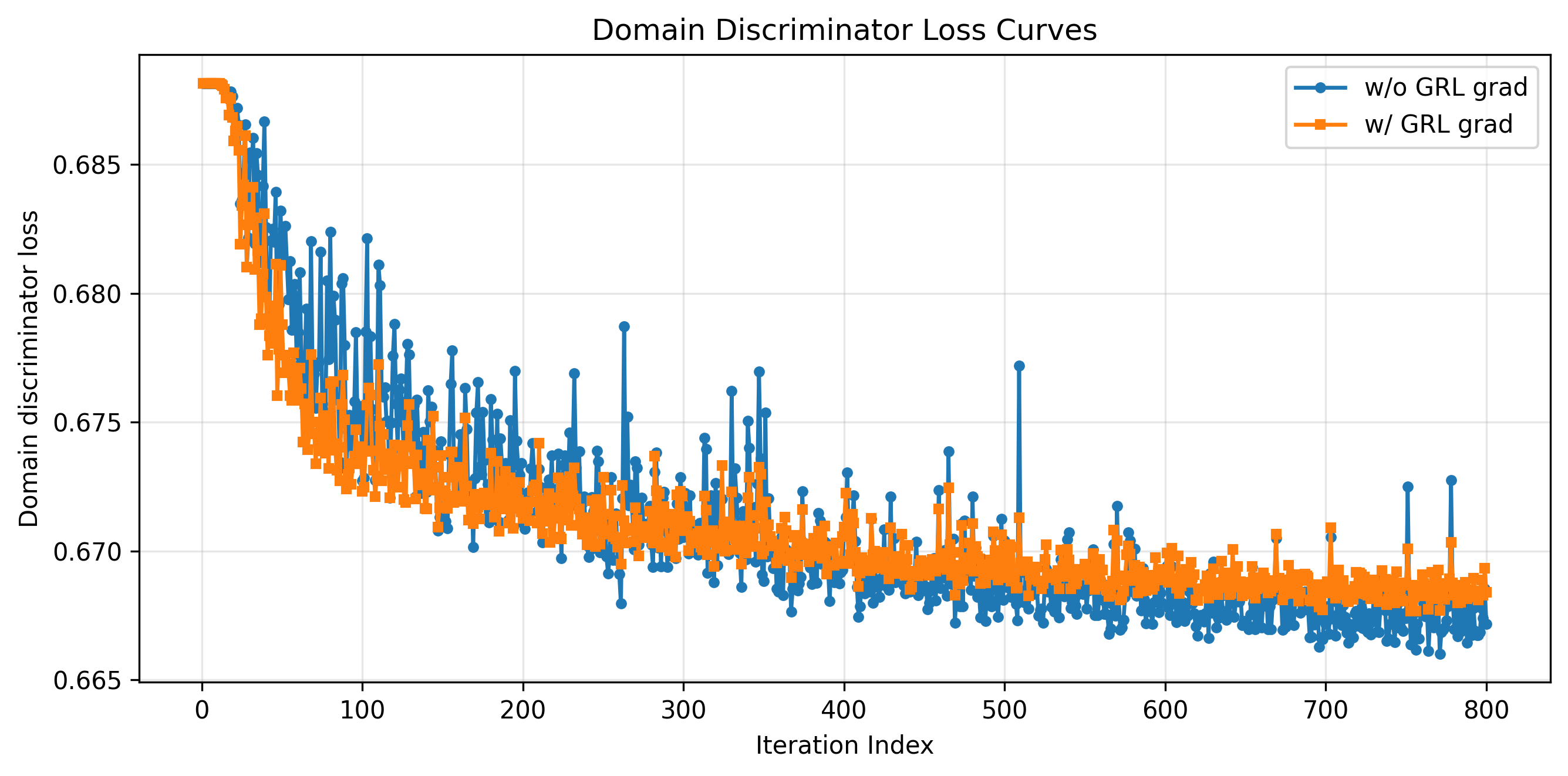}
    \caption{\color{black} Domain Discriminator Loss Curves With/Without GRL Gradients.}
    \label{figA}
 \vspace{-1em}
\end{figure}

\subsubsection{Results and observations}
\label{sec:appendix_domain_disc_results}
Fig.~\ref{figA} reports the domain discriminator loss curves with and without GRL gradients. The two curves are nearly overlapping, indicating that (i) the discriminator does not gain a clear advantage from MTN features, and (ii) enforcing domain confusion via GRL does not meaningfully change the training dynamics. Consistent with this observation, adding the discriminator/GRL does not lead to a significant difference in calibration performance in our runs.

\subsubsection{Discussion}
\label{sec:appendix_domain_disc_discussion}
These results suggest that, in practice, MTN does not rely on strongly domain-specific shortcuts even though it takes image and logit inputs. We attribute this behavior to the combined effect of: (1) \emph{complementary mixing}, which constructs samples with cross-domain regions and thereby reduces the predictiveness of domain-specific appearance/confidence cues; (2) a \emph{domain-shared} MTN, which limits capacity for domain-dependent specialization; and (3) the \emph{outcome-driven meta-objective}, which rewards MTN updates only when they improve cross-domain self-training outcomes. Together, these mechanisms are sufficient to mitigate domain-specific cue learning in our empirical study.}

\begin{figure}[ht]
    \centering
    \includegraphics[width=0.98\linewidth]{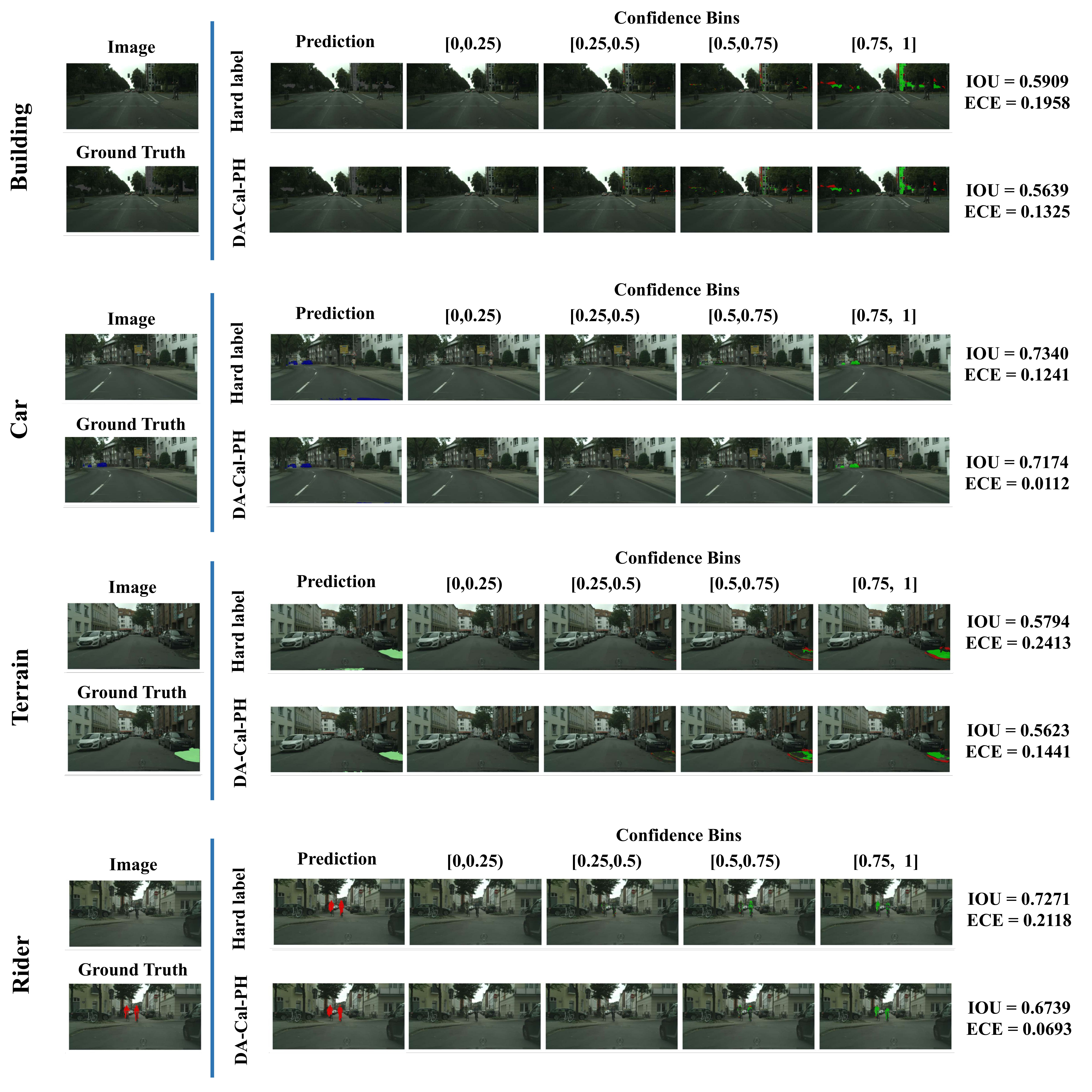}
    \caption{\color{black} \textbf{Qualitative failure cases where calibration improves but segmentation does not (Tab.~\ref{table4}: Row~1 vs Row~3).}
    We visualize four classes including frequent classes (\textit{building}, \textit{car}) and rare classes (\textit{terrain}, \textit{rider}).
    For each class, we compare the predictions produced by hard pseudo-label training (row~1) and the calibrated pseudo-label variant (DA-Cal-PH, row~3) using confidence-bin maps.
    \textbf{Green} indicates correctly segmented regions and \textbf{red} indicates incorrect regions.
    Although calibration is consistently improved (lower ECE), the IoU does not necessarily increase, illustrating that improved probabilistic calibration is not entirely equivalent to improved \(\arg\max\) segmentation accuracy in self-training-based UDA.}
    \label{fig:failure_calib_vs_miou}
     \vspace{-1em}
\end{figure}

{\color{black} 
\subsection{When does better calibration improve mIoU?}
\label{app:when_calib_helps_miou}

\textcolor{black}{Our primary contribution is a principled \emph{cross-domain calibration} method for UDA semantic segmentation, aiming to improve the reliability of predictive probabilities under domain shift. We emphasize that better calibration (probability correctness) is \emph{not} equivalent to better segmentation mIoU~\cite{guo2017calibration,wang2023calibrating}, since mIoU is mainly determined by \(\arg\max\) decisions and spatial consistency. As a result, calibration improvements should be interpreted as improving probability reliability, while mIoU gains are a secondary benefit that may (but need not) arise through better pseudo-label learning.}

\textcolor{black}{In our setting, whether calibration improvements translate into segmentation gains depends on \emph{how the probabilities are used}:}

\textcolor{black}{\noindent\textbf{(i) Calibration used for training (self-training).}
When probabilities are used to generate and/or weight pseudo supervision, better calibration can make confidence estimates more trustworthy, which typically improves the effectiveness of quality gating \(q(p)\) (e.g., confidence-based selection/weighting), reduces pseudo-label noise, and alleviates confirmation bias. However, this effect is not guaranteed to be monotonic: calibration can also make predictions more conservative (higher-entropy), which may reduce the effective pseudo-label learning signal, especially in early training when pseudo labels are noisy. Concretely, conservative probabilities may (a) decrease the number/weight of pixels or regions selected by \(q(p)\) and (b) weaken gradients from soft pseudo supervision. To mitigate this, we combine \emph{hard} pseudo labels with \emph{calibrated soft} pseudo labels: the hard branch provides a stable, low-entropy \(\arg\max\) anchor that preserves decisive supervision and improves optimization stability, while the calibrated soft branch injects uncertainty awareness and reduces overconfident confirmation bias. Under this design, mIoU improvements are more likely when improved calibration leads to cleaner and more informative pseudo labels, but they are not guaranteed to increase with calibration improvements.}

\textcolor{black}{\noindent\textbf{(ii) Calibration used only as post-hoc (inference).}
If calibration is applied purely as a post-processing step (e.g., temperature scaling at test time), it often reshapes the probability distribution without changing \(\arg\max\) decisions. In this case, calibration metrics (e.g., ECE) can improve while mIoU remains unchanged, since the predicted class label per pixel is typically identical. In some pipelines, mIoU can even slightly decrease due to interactions with thresholding-based components (e.g., confidence-based masking).}

\textcolor{black}{As a balanced perspective, Fig.~\ref{fig:failure_calib_vs_miou} shows representative cases where ECE improves while IoU slightly drops (Tab.~\ref{table4}, Row~1 vs Row~3), highlighting that better calibration (probability correctness) does not necessarily translate to better \(\arg\max\) segmentation accuracy.}
}

\vspace{1em}
{\color{black}
\subsection{\color{black} Additional Ablations on MTN Design Choices}}
\label{app:mtn_ablation}

\begin{table}[t]
\centering
\small
\setlength{\tabcolsep}{7pt}
\caption{\color{black}  Ablations of MTN design choices.}
\label{tab:mtn_ablation_gta2city_daformer}
\begin{tabular}{l|c c c c}
\hline
 & \begin{tabular}[c]{@{}c@{}}logits-only\end{tabular}
 & \begin{tabular}[c]{@{}c@{}} add \end{tabular}
 & \begin{tabular}[c]{@{}c@{}} depthwise conv\end{tabular}
 & \begin{tabular}[c]{@{}c@{}}\textbf{Ours} \end{tabular} \\
\hline
mIoU~(\(\uparrow\)) & 68.8 & 69.3 & 69.4 & \textbf{69.4} \\
ECE~(\(\downarrow\)) & 0.0791 & 0.0603 & 0.0597 & \textbf{0.0578} \\
\hline
\end{tabular}
 \vspace{-1em}
\end{table}

{\color{black}
This appendix provides additional ablations to justify MTN design choices that were fixed in the main paper. We evaluate on DAFormer under GTA$\rightarrow$Cityscapes and report mIoU (↑) and ECE (↓) in Tab.~\ref{tab:mtn_ablation_gta2city_daformer}. We focus on three factors: (i) MTN conditioning inputs (\textit{image+logits} vs.\ \textit{logits-only}), (ii) fusion operator (\textit{concat} vs.\ \textit{add}), and (iii) convolution type (\textit{standard} vs.\ \textit{depthwise}).

\paragraph{Conditioning inputs (image+logits vs.\ logits-only).}
Using \textit{logits-only} substantially worsens calibration and also reduces segmentation accuracy (mIoU 68.8, ECE 0.0791). This supports our formulation where MTN acts as a \emph{conditional translation/modulation} module that relies on the input image as a condition: image cues (texture/illumination/scene context) help the network learn how to reshape confidence in a domain-shift-aware manner. Accordingly, our default setting (\textit{image+logits/concat/standard conv}) achieves markedly better calibration while also improving mIoU (mIoU 69.4, ECE 0.0578).

\paragraph{Fusion (concat vs.\ add).}
Replacing concatenation with \textit{add} leads to a small but consistent drop (mIoU 69.3, ECE 0.0603). In our implementation, \textit{add} corresponds to copying logits to three channels and adding them to the image, which can partially entangle/lose information and reduces the capacity of the conditional mapping. In contrast, \textit{concat} preserves both signals and allows subsequent convolutions to learn an adaptive mixing, which we found slightly more effective.

\paragraph{Convolution type (standard vs.\ depthwise).}
Switching from standard to \textit{depthwise} convolution yields comparable performance (mIoU 
69.4, ECE 0.0597), indicating that our gains do not hinge on a specific convolution micro-architecture. This motivates our choice of a lightweight conv-based MTN: it offers favorable compute/memory cost and tends to be stable under noisy pseudo supervision, while maintaining competitive accuracy and calibration.

\paragraph{Connection to prior work.}
We note that our MTN is intentionally simple and can be viewed as a lightweight \emph{conditional translation/modulation} module, following established practice. Similar ``condition $\rightarrow$ modulation/translation'' designs have been validated in conditional generation/modulation and dense prediction settings, e.g., SPADE~\cite{park2019semantic}, DejaVu~\cite{borse2023dejavu}, and Local Temperature Scaling~\cite{ding2021local}. These works suggest that compact architectures are sufficient for such conditional translation modules; our ablations provide empirical evidence that the improvements are robust to reasonable MTN design choices.}

% \vfill
\bibliographystyle{splncs04}
\bibliography{main}

\end{document}